%% file: main.tex
\documentclass{article}

\usepackage{microtype}
\usepackage{graphicx}
\usepackage{subfig}
\usepackage{booktabs} 

\usepackage{hyperref}

\usepackage[accepted]{mlsys2025}

\usepackage{mathptmx} 
\usepackage[normalem]{ulem}
\usepackage{enumitem}
\usepackage{setspace}
\usepackage{nicefrac}
\usepackage[flushleft]{threeparttable}
\usepackage{booktabs}
\usepackage{multirow}
\usepackage{makecell}
\usepackage{algorithm}
\usepackage[per-mode=symbol, range-units=single, mode=text]{siunitx}

\usepackage{pifont}
\usepackage{tikz}
\usepackage{inconsolata}
\usepackage{xspace}
\usepackage[noabbrev, capitalise, nameinlink]{cleveref}

\newcommand{\design}{FastSwitch\xspace}

\newcommand{\trim}[1]{}
\newcommand{\deleted}[1]{}
\begin{document}

\twocolumn[
\mlsystitle{\design{}: Optimizing Context Switching Efficiency in Fairness-aware Large Language Model Serving}



\mlsyssetsymbol{equal}{*}
\begin{mlsysauthorlist}
\mlsysauthor{Ao Shen}{equal,pu,sqz}
\mlsysauthor{Zhiyao Li}{equal,thu}
\mlsysauthor{Mingyu Gao}{sqz,thu}
\end{mlsysauthorlist}

\mlsysaffiliation{thu}{Institute for Interdisciplinary Information Sciences, Tsinghua University, Beijing, China}
\mlsysaffiliation{pu}{Department of Computer and Information Technology, Purdue University, West Lafayette, USA}
\mlsysaffiliation{sqz}{Shanghai Qi Zhi Institute, Shanghai, China}

\mlsyscorrespondingauthor{Ao Shen}{shen634@purdue.edu}

\mlsyskeywords{Machine Learning, MLSys}

\vskip 0.3in

\input{0_abstract}
]



\printAffiliationsAndNotice{\mlsysEqualContribution} 

\input{1_introduction}
\input{2_background_and_motivation}
\input{3_design}

\input{4_methodology}
\input{5_evaluation}

\input{6_related_work}
\input{7_conclusion}

\bibliography{refs}
\bibliographystyle{mlsys2025}

\appendix

\end{document}

%% file: 0_abstract.tex
\begin{abstract}
Serving numerous users and requests concurrently requires good fairness in Large Language Models (LLMs) serving system. This ensures that, at the same cost, the system can meet the Service Level Objectives (SLOs) of more users , such as time to first token (TTFT) and time between tokens (TBT), rather than allowing a few users to experience performance far exceeding the SLOs. To achieve better fairness, the preemption-based scheduling policy dynamically adjusts the priority of each request to maintain balance during runtime.  However, existing systems tend to overly prioritize throughput, overlooking the overhead caused by preemption-induced context switching, which is crucial for maintaining fairness through priority adjustments. In this work, we identify three main challenges that result in this overhead. 1) Inadequate I/O utilization. 2) GPU idleness. 3) Unnecessary I/O transmission during multi-turn conversations. Our key insight is that the block-based KV cache memory policy in existing systems, while achieving near-zero memory waste, leads to discontinuity and insufficient granularity in the KV cache memory. To respond, we introduce \design{}, a fairness-aware serving system that not only aligns with existing KV cache memory allocation policy but also mitigates context switching overhead. Our evaluation shows that \design{} outperforms the state-of-the-art LLM serving system vLLM with speedups of 1.4-11.2$\times$ across different tail TTFT and TBT.
\end{abstract}

%% file: 1_introduction.tex
\section{Introduction}
Large Language Models (LLMs) like GPT-3~\cite{chatgp3_arxiv2023}, LLaMA~\cite{llama_arxiv2023}, and Qwen~\cite{qwen2_arxiv2024} have revolutionized AI by powering applications such as language translation, conversational agents, and code generation~\cite{codegen_arxiv2023, memory_arxiv2024, translation_arxiv2023, measuring_arxiv2020, large_arxiv2024, frage_arxiv2018, piqa_aaai20, chatgpt}, driving the adoption of Model-as-a-Service (MaaS) platforms~\cite{sglang_arxiv2023, vllm_sosp23, flexgen_icml23, agrawal_arxiv2023, sarathi_arxiv2023, deepspeed_arxiv2022}.
However, serving LLMs is highly expensive, typically requiring a significant number of hardware accelerators, such as GPUs with High-Bandwidth Memory (HBM)~\cite{hbm_arxiv2021}.
To efficiently process numerous inference requests, transformer-based LLM serving typically requires storing the key-value cache (KV cache) of requests in GPU memory. While GPUs offer HBM, which significantly boosts I/O performance, this memory is both expensive and limited in capacity. Moreover, advanced features such as Chain-of-Thought (CoT) reasoning and multimodal inputs~\cite{chain_arxiv2022, multi_nips24} amplify these constraints by demanding longer context lengths.

To ensure service quality, LLM applications assign Service Level Objectives (SLOs) to users. Given a fixed serving cost, a key insight is that, to serve more requests and users at one time, it is preferable to leverage limited resources to meet the SLOs of as many users as possible, rather than allowing a few users to experience performance far exceeding their SLOs. Therefore, fairness in serving becomes a critical concern. Many LLM serving systems achieve fairness by dynamically adjusting request priorities during runtime based on SLOs. Given the constant constraints on GPU memory, the preemption mechanism is essential for terminating lower-priority requests, efficiently managing resources, and ensuring fairness. And most existing systems~\cite{vllm_sosp23, sglang_arxiv2023} adopt swapping as the default preemption approach, each preemption results in context switching. In our work, context switching refers to requests and their KV cache being swapped between GPU and CPU memory based on their priorities. Recent works such as VTC~\cite{fairness_osdi24}, Andes~\cite{andes_arxiv2024}, QLM~\cite{qlm_arxiv2024}, FastServe~\cite{fastserve_arxiv23}, and LLMS~\cite{llmaas_arxiv2024} have explored various strategies for preemptive scheduling to ensure fairness and mitigate issues like head-of-line blocking in LLM serving systems. Furthermore, in multi-turn conversations, preemption involves transferring the KV cache of completed conversations to CPU memory for delayed future use. Finally, the decode-prefill disaggregation architecture~\cite{dist_arxiv2024} also introduces important applications for preemption and KV cache transfer between tasks.

However, existing serving systems focus too heavily on throughput, overlooking the priority adjustment overhead required to meet the SLOs of more users.
In existing system~\cite{vllm_sosp23}, the non-contiguous paged memory for KV cache to achieve higher throughput causes fragmented memory allocation, leading to poor utilization of PCIe I/O bandwidth when context switching. Additionally, the current scheduling design causes preemption to stall inference, leaving the GPU idle. What's more, when serving multi-turn conversations, the redundant swap out volume of previous conversations KV cache generates unnecessary I/O transfers, wasting bandwidth. These challenges introduce significant overhead and degrade key SLOs such as TTFT and TBT, ultimately reducing Quality-of-Experience (QoE)~\cite{andes_arxiv2024}. As shown in \cref{fig:context_switch_overhead}, the stall time caused by preemption can be several times longer than the inference time of a single iteration.

Addressing the overhead from frequent context switching is crucial for improving the fairness and scalability of LLM serving systems. Recent works have improved the overhead in many ways. However, our detailed characterization reveals three challenges that remain systematically unaddressed. First, to tackle I/O inefficiency, previous approaches, such as Llumnix~\cite{llumnix_arxiv2024}, use an additional buffer to first merge the KV cache before performing the transfer. However, we found out that it failed to fully utilize bandwidth due to limited granularity and added design complexity with the overhead of a second transfer for data merging. Second, to avoid GPU idleness, AttentionStore supports layer-wise asynchronous swapping, but they interfere with CUDA graph execution during inference, increasing inference time, especially when swapping latency exceeds inference time. The static nature of CUDA graphs, makes it difficult to integrate swapping subgraphs seamlessly. FastServe adopts iteration-wise transmission to predict and pre-load KV cache while supporting inference CUDA graph execution for GPU efficiency. However, predicting suitable requests for preemptive swap out is not easy. Also, it fails to achieve asynchronous dispatch between swapping APIs and inference kernels. Most importantly, they both failed to identify that the bottleneck in swapping is the overhead of the API dispatch stage. Finally, to reduce unnecessary I/O usage, AttentionStore~\cite{attentionstore_arxiv2024} explored prefix reuse in multi-turn conversations by offloading the KV cache to multi-tier storage after each sub-conversation, preserving previous turn contexts and reducing swap out volume by swapping out only the KV cache of the new turn. However, in GPU-CPU memory configurations, the limited capacity of CPU memory introduces a new challenge: it becomes difficult to fully utilize KV cache copies. High-priority requests can invalidate lower-priority backups and complicating the creation of complete prefill copies for subsequent turns. The core challenge is finding an effective way to reuse the previous KV cache during preemption.

In response, we introduce \design{}, a new serving system that optimizes preemptive context switching in LLM inference with minimal additional cost. \design{} leverages an I/O-aware KV cache management strategy, the Dynamic Block Group Manager, which enhances bandwidth utilization and reduces kernel dispatch overhead by allocating memory for KV cache at a coarser granularity. Additionally, \design{} integrates a Multithreading Swap Manager to asynchronously manage KV cache transfers, minimizing delays from cache dependencies during context switching and improving token generation efficiency. Finally, the KV Cache Reuse Mechanism, integrated into the Dynamic Block Group Manager, manages and reuses KV cache copies across multi-turn dialogues, thereby reducing unnecessary I/O usage. As a result, \design{} effectively addresses the overhead from frequent context switching caused by high-frequency priority changes, ensuring fairness and responsiveness in multi-users environments while maintaining efficient resource utilization.
In summary, this paper makes the following key contributions:

\begin{itemize}[leftmargin=1.2em, itemsep=0pt, topsep=0pt]
    \item We characterize and identify three major challenges—I/O underutilization, GPU idleness, and redundant I/O usage—that hinder fairness and overall SLO improvements when adjusting priority, yet remain unresolved in prior works.  
    \item We propose a fairness-aware serving system, \design{}, which enables efficient preemptive context switching by incorporating three key optimizations that work synergistically to address these challenges in LLM serving.
    \item Our detailed evaluation demonstrates substantial improvements compared with the state-of-the-art serving system, particularly under high priority-update frequency. Specifically, we achieve a speedup in TTFT of up to 1.4-5.8$\times$, in TBT of up to 11.2$\times$, and an increase in Throughput of up to 1.44$\times$.
\end{itemize}

%% file: 2_background_and_motivation.tex
\section{Background and Motivation}
\label{sec:background}
\subsection{Preemption-based Scheduling in LLM Inference}
\label{subsec:preemption_scheduling}
Preemptive scheduling ensures resource prioritization under constrained HBM. There are two main preemption methods: recomputation, which halts a task and recomputes its KV cache upon resumption, increasing latency and computation resource use, especially for long contexts; and swapping, which transfers the KV cache between GPU and CPU memory, avoiding recomputation. Systems like vLLM use swapping to optimize memory usage. Recent works explore various preemptive scheduling strategies: Andes~\cite{andes_arxiv2024}, FASTSERVE~\cite{fastserve_arxiv23}, and QLM~\cite{qlm_arxiv2024}. They focus on mitigating head-of-line blocking and considering the SLOs when scheduling. Additionally, LLMS~\cite{llmaas_arxiv2024} manages multiple LLM contexts across apps. Our design builds on vLLM by optimizing its swapping efficiency.

\subsection{Motivation}
\label{subsec:motivation}
\begin{figure}[htbp]
    \centering
    \begin{minipage}[t]{0.48\linewidth}
        \centering
        \includegraphics[width=0.95\linewidth]{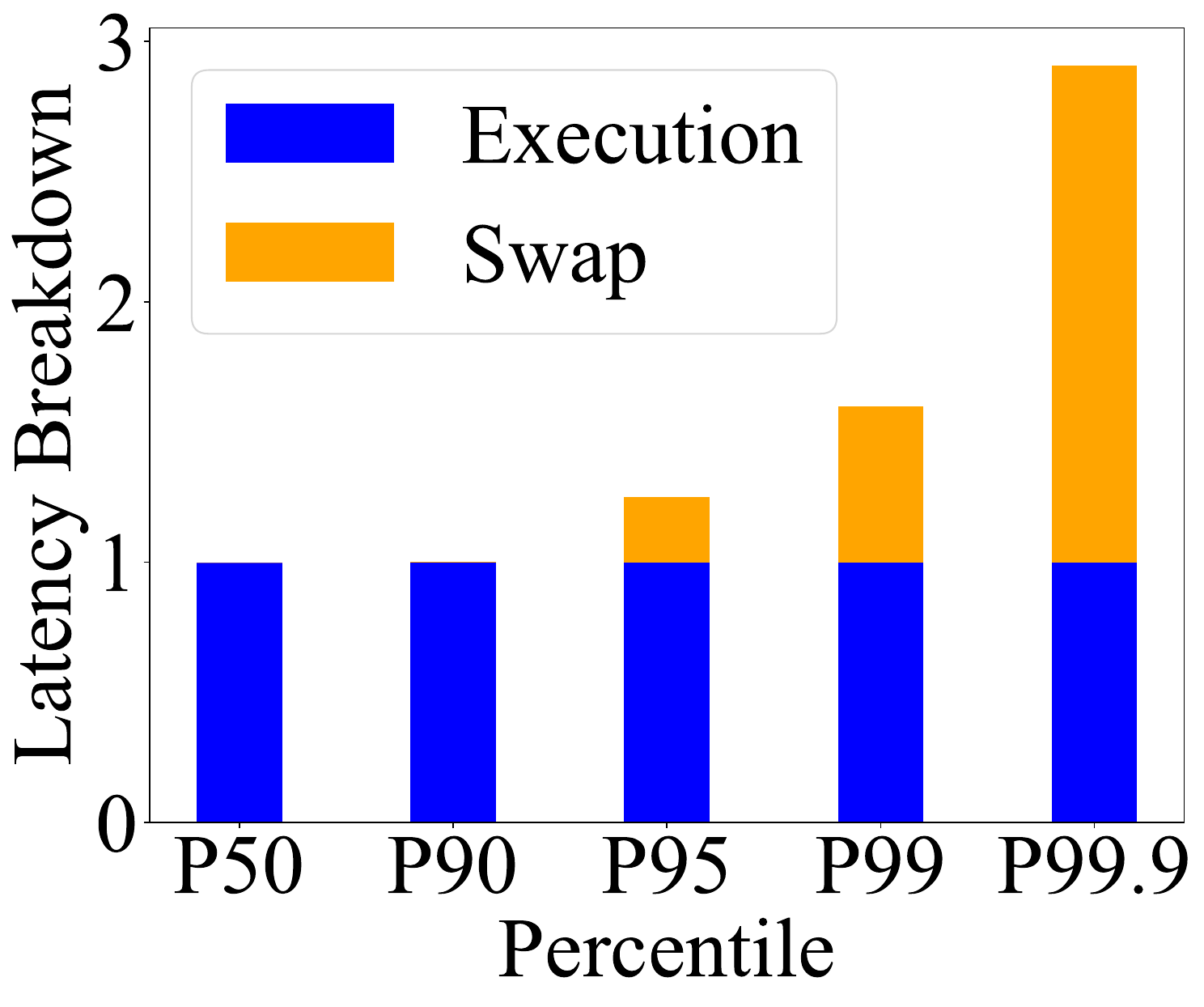}
        \caption{Latency breakdown across percentiles.}
        \label{fig:context_switch_overhead}
    \end{minipage}
    \hspace{0.01\linewidth} 
    \begin{minipage}[t]{0.48\linewidth}
        \centering
        \includegraphics[width=1.0\linewidth]{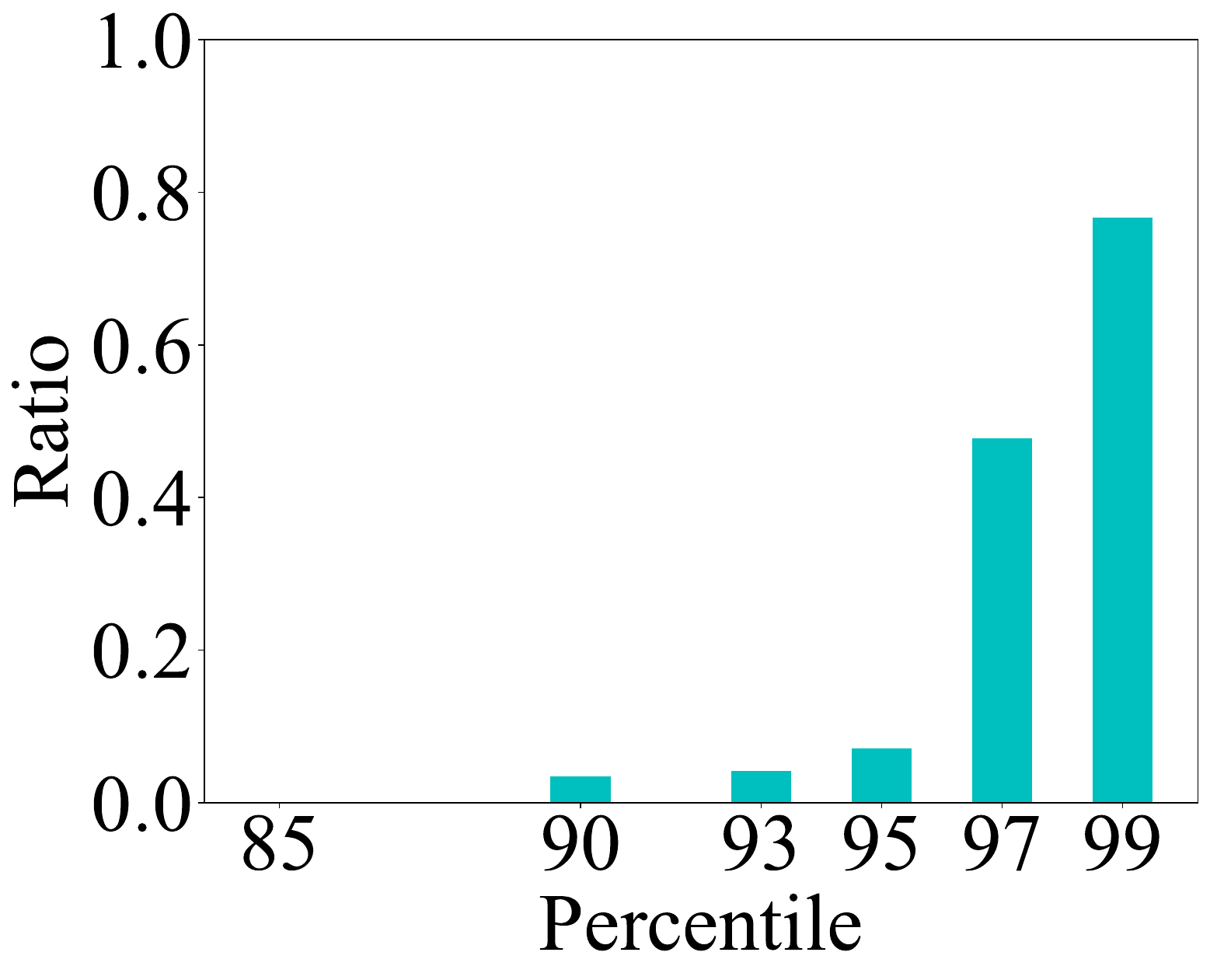}
        \caption{Only a small proportion of requests need to wait for the KV cache in most iterations.}
        \label{fig:num_swap_versus_running}
    \end{minipage}
\end{figure}
\textbf{Observation: Context Switching Overhead.} 
Preemptive context switching, necessary for managing dynamic workloads and prioritizing requests, introduces significant overheads, particularly from KV cache I/O operations. These affect SLOs like TTFT and TBT. The impact worsens with longer contexts and increased preemption overhead.

An experiment using the LLaMA-8B model served with vLLM on an A10 24GB GPU involved processing 1,000 multi-turn requests from the ShareGPT dataset with average request rate of 1 req/s and priority updates every 100 iterations. We normalized the latency by setting the execution time of inference to 1. The latency penalty from preemption was measured as KV cache swapping time.

\cref{fig:context_switch_overhead} shows that P99 latency is approximately 1.6 times higher than the P50, with swapping-induced stall time accounting for about 59.9\% of P99 latency. This reveals significant performance degradation in high-stress scenarios, which becomes even more pronounced at P99.9 where the total latency increases to nearly twice the inference time.

\begin{figure}[ht]
    \centering
    \includegraphics[width=0.8\linewidth]{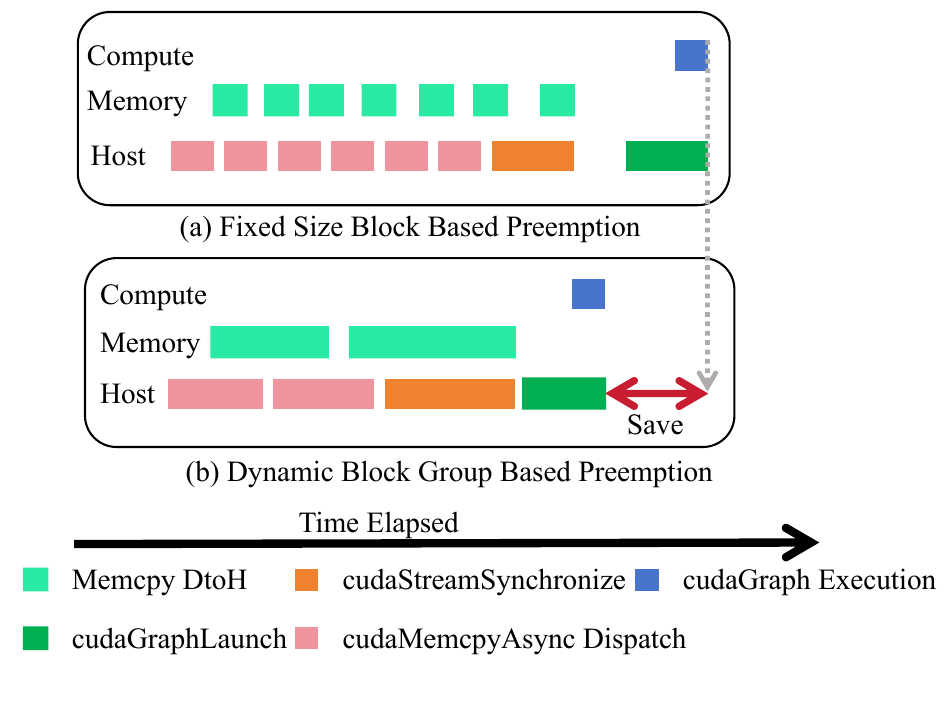}
    \caption{Timeline comparison of fixed-size block based preemption and dynamic block group based preemption.}
    \label{fig:bottleneck_when_dispatch_blocks}
\end{figure}
\textbf{Challenge \#1: Low Bandwidth Utilization.}
While vLLM’s KV cache management policy effectively reduces internal fragmentation by using non-contiguous virtual memory for the KV cache, optimizing swapping of the KV cache remains a significant challenge. As illustrated in \cref{fig:bottleneck_when_dispatch_blocks}(a), performance bottlenecks arise due to suboptimal KV cache swapping granularity, such as small 128 KB KV cache swapping granularity in LLaMA-8B. The dispatch overhead for each \texttt{cudaMemcpyAsync} call exceeds its \SI{10}{\micro\second} execution time, leading to idle I/O. This issue is further exacerbated by the fact that the transfer size is below PCIe 4.0’s optimal 320 KB, thus reducing efficiency. In this setting, dispatch time accounts for 90\%-95\% of the total transmission time.

Previous works such as Llumnix~\cite{llumnix_arxiv2024} tackle bandwidth inefficiencies caused by insufficient granularity in I/O management. They attempt to increase I/O utilization an additional small buffer, but their approach is still unable to fully exploit the available bandwidth because of limited granularity. Moreover, merging to buffer introduces additional design complexity and overhead due to the necessity of a second transfer.

Furthermore, simply increasing the block size in vLLM could cause internal fragmentation. And it would undermine the memory utilization efficiency that vLLM strives to achieve. To address this, vLLM sets the default block size to 16 tokens, aiming to strike a balance between memory efficiency and performance. Traditional approaches that preallocate KV cache memory increase fragmentation, while dynamic allocation introduces complexity and overhead during context switching. Therefore, developing lightweight, adaptive memory management strategies that align with vLLM’s policies is critical for maintaining both efficient memory usage and larger transfer granularity to better utilize bandwidth. This remains a key challenge that needs to be addressed.


\textbf{Challenge \#2: GPU Idleness During Preemption.} 
We evaluated LLaMA-8B on an NVIDIA A10 GPU using a Markov priority pattern with priority-update frequency = 0.02 and 500 multi-turn conversations from ShareGPT. As shown in the \cref{fig:num_swap_versus_running}, the impact of global priority updates across requests is most pronounced in tail cases, where a significant proportion of requests experience delays. In most scenarios, KV cache transfers between CPU and GPU memory affect only a subset of requests. Although most requests proceed without interruption, the overhead from preemption can exceed a single inference step. This leads to inference stalls, resulting in GPU being idle.

To address GPU idleness, previous work like AttentionStore~\cite{attentionstore_arxiv2024} supports layer-wise asynchronous swapping. However, this approach can interfere with the execution of CUDA graphs during inference and increase inference time, especially when swapping latency exceeds inference time. The core challenge lies in the nature of CUDA graphs, which are generated through static compilation. If one attempts to integrate swapping subgraphs into an existing execution graph, given that batch size is a parameter of the node in graph, it becomes necessary to iterate over all possible batch size values across both the swapping subgraph and the execution subgraph. This requires a significant amount of GPU memory for the storage of CUDA graphs because each CUDA graph needs to reserved space for input, activations, and output. Also, profiling a new graph on-the-fly every time is impractical. On the other hand, if integration is not performed, the static execution graph cannot dynamically adjust its size to accommodate temporary changes. For these reasons, the latest version of vLLM has deprecated this mechanism. Moreover, without an I/O-aware KV cache allocator to increase transfer granularity, the bottleneck of swapping lies in the overhead of the \texttt{cudaMemcpyAsync} dispatch stage rather than the execution stage as mentioned in \textbf{Challenge \#1}. This issue cannot be resolved by layer-wise swapping.

FastServe~\cite{fastserve_arxiv23} adopts iteration-wise transmission to predict and pre-load KV cache while supporting inference CUDA graph execution for GPU efficiency. However, predicting suitable requests for preemptive swap out can be challenging. Additionally, it does not achieve asynchronous dispatch between swapping APIs and inference kernels, thus failing to resolve the major part of overhead as stated in \textbf{Challenge \#1} either. It also overlooks the difficulty of maintaining a coherent order of swapping dispatches, which will be discussed in \cref{design:mt_swap_manager}.


\begin{figure}[ht]
    \centering
    \includegraphics[width=\linewidth]{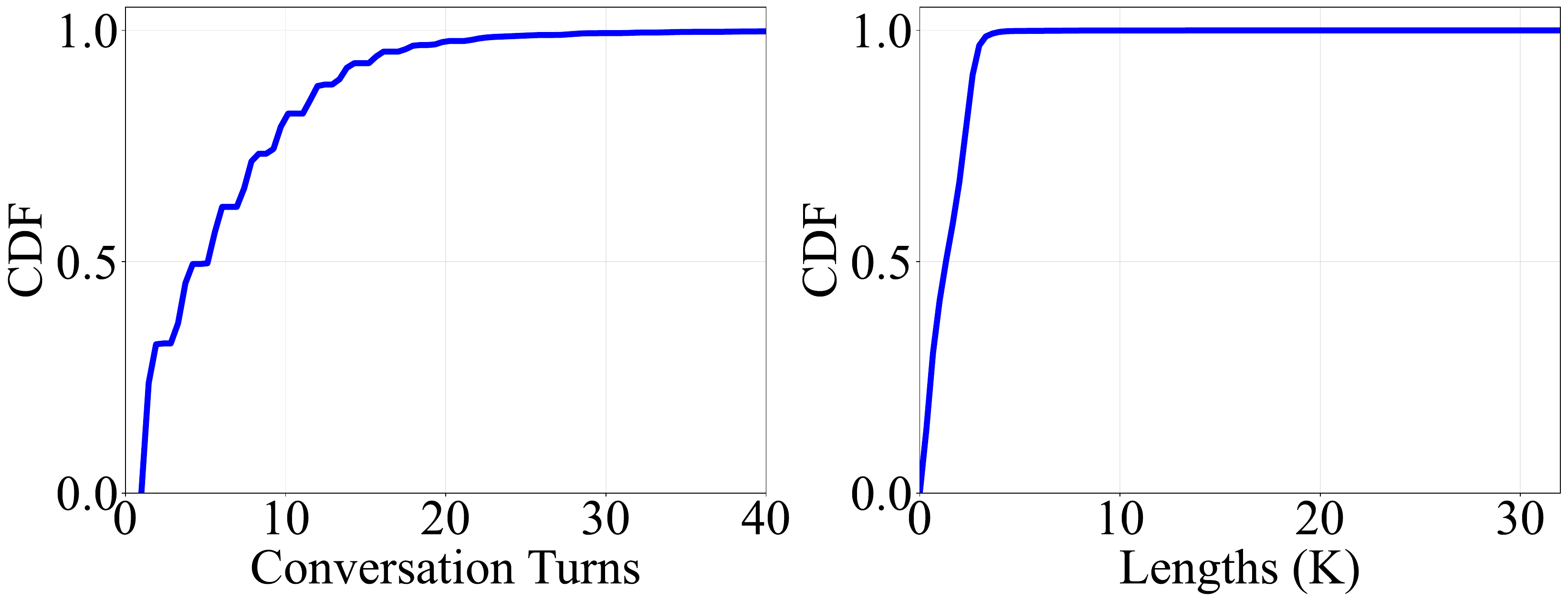}
    \caption{ShareGPT conversation turns \& lengths distribution.}
    \label{fig:sharegpt_distributions_cdf}
\end{figure}

\textbf{Challenge \#3: Contaminated CPU KV Cache Copies in Multi-turn Conversations.}
Multi-turn conversations dominate real-world LLM applications like chatbots. Datasets like ShareGPT (~100K conversations) show 78\% of interactions involve multiple turns, averaging 5.5 turns per conversation. These require maintaining large KV caches for contextual coherence, but repetitive patterns cause KV cache recomputation, leading to inefficiencies. AttentionStore~\cite{attentionstore_arxiv2024} tackles redundant KV cache recomputation by reusing prefixes in multi-turn conversations. It stores the full KV cache in multi-tier storage after each conversation and swaps out only the incremental KV cache for new turns, as previous context is already preserved, reducing swap volume.
However, in the hardware setting where memory resources for KV cache are solely allocated to GPU and CPU memory, such an approach becomes challenging. Not all requests can fully utilize KV cache copies stored in CPU memory given that CPU memory is not unlimited. Intuitively, when a high-priority request requires memory allocation, the system prioritizes reclaiming memory from lower-priority requests, which invalidates lower-priority requests' KV cache backups. This presents a new challenge: to create a complete copy for prefill the next turn with prefix, how we can minimize the unnecessary removal of the previous context when the KV cache memory is preempted by other tasks.


%% file: 3_design.tex
\section{Design of \design{}}
\label{sec:design}
\begin{figure}[ht]
    \centering
    \includegraphics[width=\linewidth]{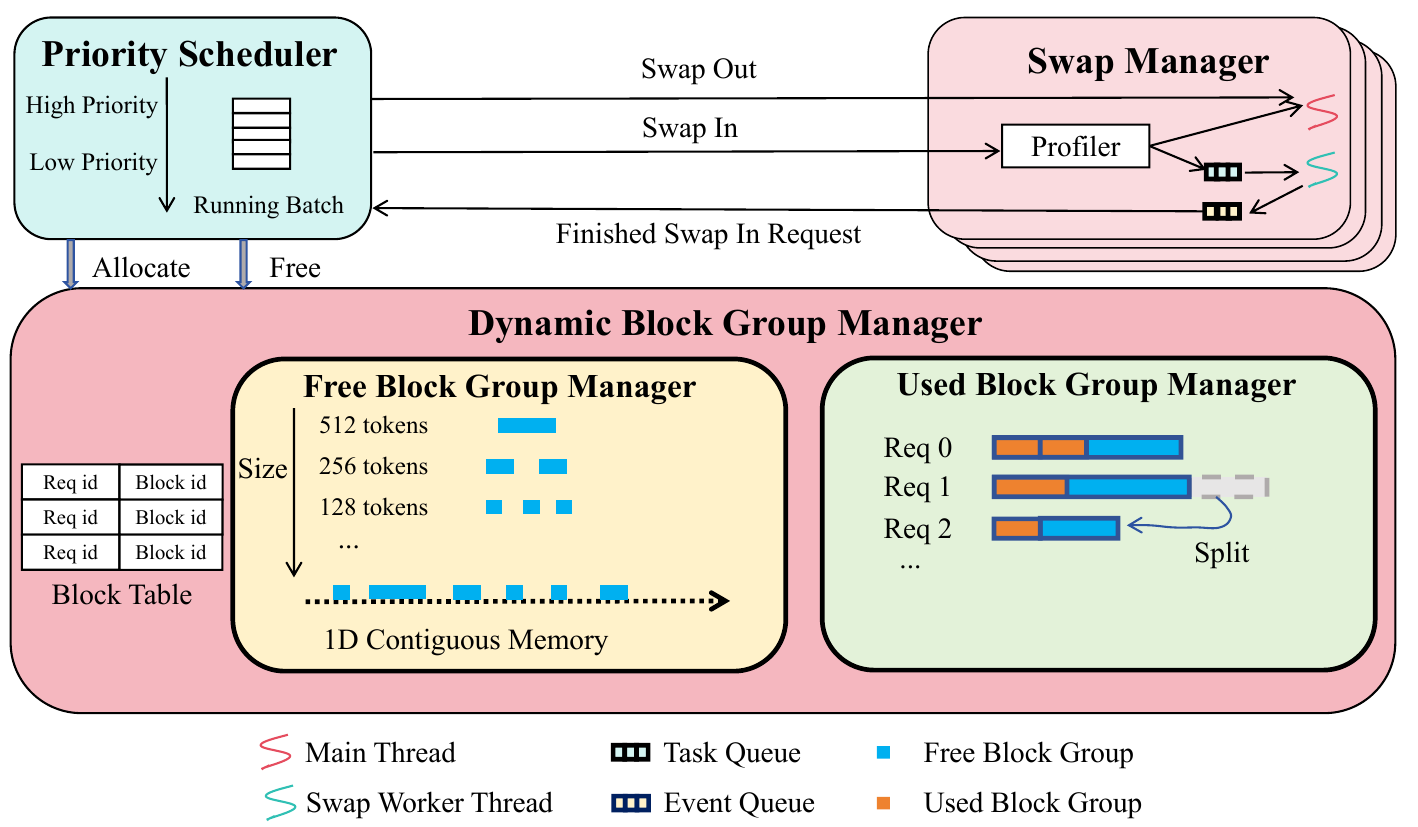}
    \caption{\design{} system overview.}
    \label{fig:system_architecture}
\end{figure}

The architecture of \design{} is shown in \cref{fig:system_architecture}. \design{} incorporates three key optimizations that work collaboratively to tackle the challenges discussed in \cref{subsec:motivation}. First, \design{} adopts a \textbf{Priority Scheduler} to schedule high-priority requests into the running batch based on the latest priorities. Next, to address \textbf{Challenge \#1}, the \textbf{Dynamic Block Group Manager} in \cref{design:dynamic_bg_manager} handles instructions of KV cache allocation from the scheduler. Based on demand and availability, it prioritizes fulfilling requests from the KV cache in the Free Block Group Manager first and then from the Used Block Group Manager if necessary, thus increases the bandwidth utilization through larger transfer granularity. 
Then, building on this foundation, the \textbf{Multithreading Swap Manager} in \cref{design:mt_swap_manager} leverages async swapping to avoid the GPU idleness in \textbf{Challenge \#2}. When requests need to be swapped in, the profiler determines whether to perform asynchronous swapping. If asynchronous swapping is chosen, the swapping task is added to the task queue, and worker threads from the thread pool will then execute the task. These threads record progress using CUDA events retrieved sequentially from the event pool. At the beginning of each iteration's scheduling phase, the scheduler retrieves completed requests from the swap manager and returns them to the running batch. If asynchronous swapping is not used, the main thread will then handle the operation, stalling the inference process and waiting for the swap in to complete.
Finally, this manager tracks block group usage and integrates the \textbf{KV Cache Reuse Mechanism} discussed in \cref{design:kv_cache_reuse}. When requests are swapped out from the GPU to the CPU, the scheduler first interacts with the manager to determine how many KV cache copies in the CPU can be reused, thereby minimizing the volume of KV cache transfers discussed in \textbf{Challenge \#3}.

\subsection{Dynamic Block Group Manager for Increased Granularity and I/O Bandwidth Utilization}
\label{design:dynamic_bg_manager}
In this section, to address \textbf{Challenge \#1},
we introduce the Dynamic Block Group Manager, an I/O-aware KV cache allocator. Instead of managing individual blocks, this manager handles multiple blocks simultaneously, reducing dispatch time and improving I/O bandwidth utilization. By leveraging the principles of buddy allocation, the Dynamic Block Group Manager aims to allocate memory blocks that closely match the size required by each request, thereby optimizing transfer efficiency. The Dynamic Block Group Manager is designed to be pluggable into existing systems. It seamlessly integrates with vLLM’s KV cache management policy, maintaining high throughput while improving I/O utilization during preemption.

\textbf{Dynamic Block Group Allocation and Management.}  
As depicted in \cref{fig:system_architecture}, the key idea behind Dynamic Block Group Manager, is analogous to the buddy allocator~\cite{buddy_1975} used in operating systems. The KV cache memory is allocated in larger chunks referred to as block groups, each comprising multiple contiguous vLLM blocks. Each request is assigned one or more block groups to store its KV cache. The most recently allocated block group for a request is considered active. This active block group not only contains new KV cache for the current request but can also be split into one or more smaller block groups to serve other requests. We first perform block/page-based preallocation similar to vLLM. And then merge all blocks into an initial free block group. Based on the need of either size or address, this free block group is subsequently split into two or three smaller groups. The Dynamic Block Group Manager organizes block groups through two primary subcomponents: the Free Block Group Manager and the Used Block Group Manager. 

To maintain optimal memory usage, the manager supports dynamic splitting and merging of block groups. 
When no matching free block group is available in the Free Block Group Manager, the active block group currently being used by a randomly selected request can be taken from the Used Block Group Manager. The manager can then split either a portion of the free space or all of it to fulfill the needs of another request, as illustrated in \cref{fig:system_architecture}. The unused portions are then reallocated to accommodate other requests as needed. Conversely, if multiple adjacent free block groups are available, they can be merged to form larger block groups. This merging enhances memory continuity and further reduces fragmentation. We initially set the expected size of the first block group for each request to 60 blocks, which corresponds to approximately 1,000 tokens when the block size is 16 tokens. The manager dynamically adjusts this size to meet the expected KV cache requirement of each request, taking into account the current availability of free KV cache.

\textbf{Bandwidth Utilization Improvement.}  
The Dynamic Block Group Manager enables larger granularity transfers by managing memory at the block group level, reducing the number of transfers and eliminating associated latency. As shown in \cref{fig:bottleneck_when_dispatch_blocks} (b), compared to fixed-size blocks, our design consolidates smaller memory operations into fewer, larger transfers, thereby reducing dispatch overhead and improving PCIe utilization. Llumnix introduces an additional 2-block buffer to merge the KV cache before performing a secondary transfer. However, this granularity is insufficient to fully utilize I/O bandwidth and the extra transfer brings complexity and additional latency. Moreover, simply increasing the buffer size risks excessive space usage, which contradicts vLLM's goal of minimizing waste through on-demand allocation. Our approach not only improves I/O utilization by providing better granularity but also avoids the need for secondary transfer. For example, when deploying LLaMa-8B on an NVIDIA A10 GPU, our method achieves an average granularity of approximately 20 blocks per block group. This result is observed across seven distinct frequencies, with a median priority-update frequency of 0.02 (once every 50 iterations). This maximizes bandwidth efficiency while avoiding the complication associated with secondary transfer.

\subsection{Multithreading Swap Manager for Optimizing Token Generation Efficiency}
\label{design:mt_swap_manager}
\begin{figure}[ht]
    \centering
    \includegraphics[width=\linewidth]{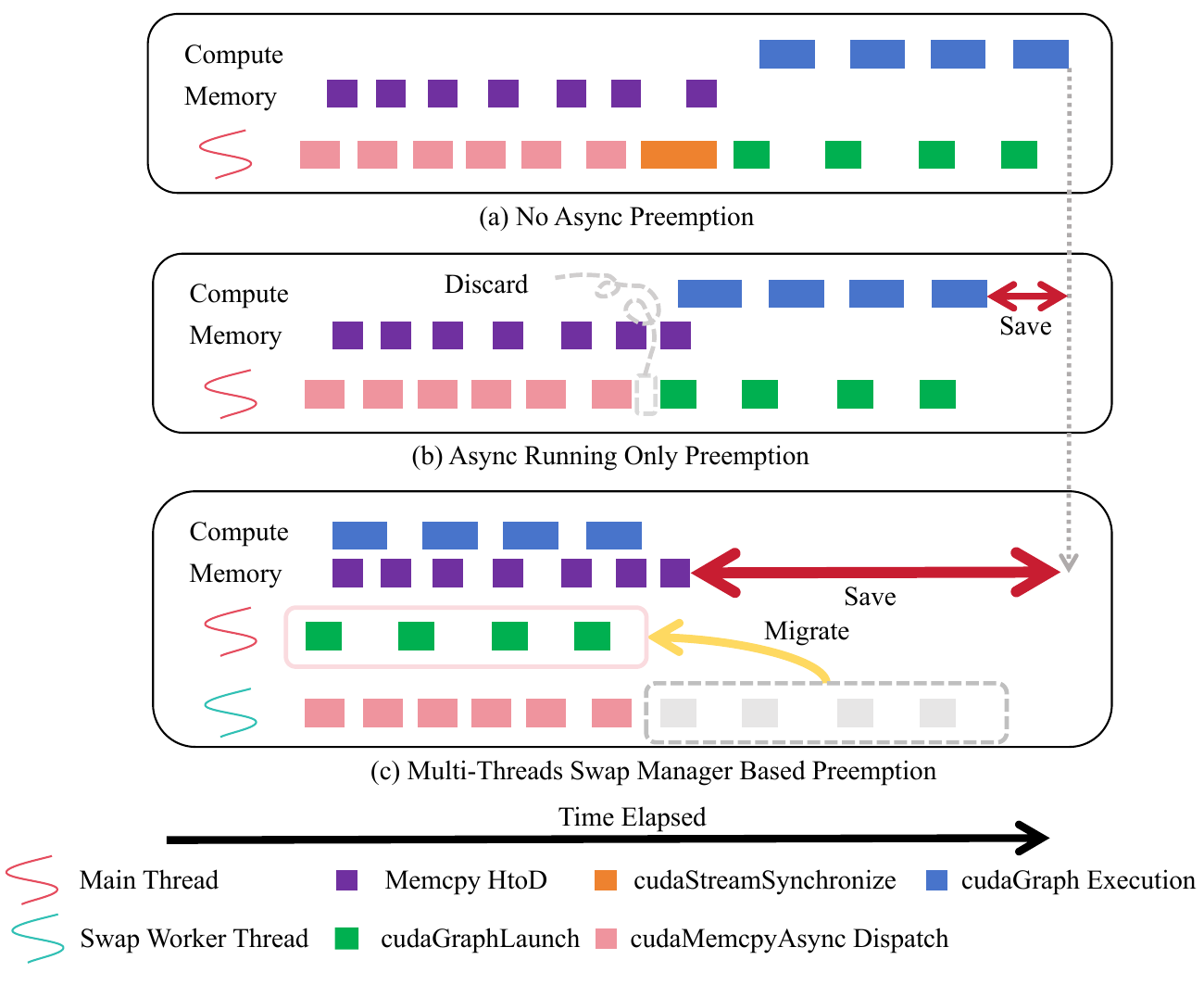}
    \caption{Comparison of varying degrees of asynchronous preemption.}
    \label{fig:async_result}
\end{figure}

Built upon the Dynamic Block Group Manager from the first section, we now introduce the Multithreading Swap Manager to comprehensively address \textbf{Challenge \#2} in an iteration-wise manner.

\textbf{Adaptive Swapping Strategy.}  
 As shown in \cref{fig:system_architecture}, the Swap Manager employs a dynamic swapping strategy based on the system's current state. To enable informed decision-making, a profiler monitors key metrics such as the number and size of ongoing swapping operations over a recent time window. We observe that asynchronous handling of preemption is not always the optimal solution. While asynchronous swapping in generally reduces idle time and improves efficiency, it's not always the best approach. Specifically, when the total number of requests is high, but each request is relatively short, asynchronous swap in may degrade token generation efficiency. In such cases, it is more beneficial to perform synchronous swap in, as the overhead of swapping is minimal compared to the potential gain from processing a larger number of tokens. This adaptive strategy carefully balances the swapping overhead with token generation efficiency, dynamically switching between asynchronous and synchronous swap in based on workload run time metrics.

\textbf{Overcoming Python GIL Limitation and Ensuring Conflict-free Dispatch Order of Multi-stream CUDA Runtime APIs.}  
We observed that Python-based call stacks in many serving systems introduce the Global Interpreter Lock (GIL), which bottlenecks parallel execution of asynchronous tasks by limiting CPU-side kernel dispatch or API dispatch. While the execution stage remains unaffected, the restricted dispatch reduces the benefits of asynchronous optimizations and constrains overall system throughput and scalability. To mitigate this, we offloaded the dispatch process to C++, where we created a thread pool and used worker threads to dispatch the APIs and create CUDA events, thereby gaining fine-grained control over the entire process.

We also identified another implementation challenge: In the same CUDA context, the dispatch order of CUDA memcpy APIs across multiple streams needs to manage in a proper way. The key insight here is that in the execution stream, \texttt{cudaMemcpyAsync} calls will also be issued. If the swapping stream has already dispatched a large number of \texttt{cudaMemcpyAsync} operations, even if the inference stream has higher priority, it cannot preempt the execution stage of \texttt{cudaMemcpyAsync} in the swapping stream. This results in inference stalls and GPU being idle because the \texttt{cudaMemcpyAsync} in the inference stream must wait for I/O resource to become available in order to complete. To address this issue, we implemented fine-grained synchronization control. After a certain number of dispatches, we perform synchronization to ensure that high-priority APIs can be inserted into the dispatch queue and dispatched successfully. Although this introduces a small synchronization overhead, it is insignificant compared to the performance gains achieved through overlap.

\textbf{KV Cache Conflict Resolution in Asynchronous Swapping.} 
While asynchronous KV cache transfers enable the overlap of swapping and inference operations, boosting token generation efficiency, they also introduce the risk of KV cache conflicts between the KV cache of ongoing swapping requests and newly allocated KV cache from running requests. To address this issue, we leverage the Dynamic Block Group Manager to monitor block group allocation and usage. When KV cache conflict is detected, the Swap Manager synchronizes KV cache transfer event in a fine-grained manner, minimizing resource contention and resolving KV cache conflicts. This ensures efficient operation during overlapping swapping and inference.

\textbf{GPU Utilization Improvement.}  
By allowing certain swapping operations to occur in parallel with active inference processes, the Swap Manager reduces idle time and enhances overall throughput. This asynchronous handling ensures that the majority of requests can proceed without being delayed by swapping dependencies. As shown in \cref{fig:async_result}, in (a), no asynchronous preemption is applied, meaning all operations, such as memory copies (\texttt{Memcpy HtoD}), \texttt{cudaGraphLaunch}, and \texttt{cudaGraph} execution, are performed sequentially. This leads to inefficient resource utilization, as tasks are serialized, resulting in longer overall inference time.

In (b), only the execution stage of \texttt{cudaMemcpyAsync} is asynchronous, while the \texttt{cudaMemcpyAsync} dispatch remains synchronous. This limits the efficiency gains as the dispatch phase still causes delays and prevents full concurrency.

In contrast, (c) implements a fully asynchronous approach, where both the dispatch and execution stage of \texttt{cudaMemcpyAsync} are asynchronous to the inference. This allows for improved concurrency and more efficient resource utilization, leading to better overall performance.

\begin{algorithm}[tb]
    \caption{Multithreading Swap Manager Algorithm}
    \label{alg:multi_threaded_swap_manager}
    \footnotesize
\begin{algorithmic}
    \STATE \textbf{Initialization:}
    \STATE \quad \texttt{running} $\gets$ InitializeQueue()
    \STATE \quad \texttt{swapped} $\gets$ InitializeQueue()
    \STATE \quad \texttt{ongoing\_swap\_in} $\gets$ InitializeQueue()
    \STATE \textit{\# Recent Swapping Information}
    \STATE \quad \texttt{r\_info} $\gets$ InitializeQueue() 
    
    \FOR{each iteration}
        \STATE \textit{\# Step 1: Verify swap In Completion}
        \FOR{each $req$ in \texttt{ongoing\_swap\_in}}
            \IF{IsCompleted($req$)}
                \STATE Move($req$, \texttt{ongoing\_swap\_in}, \texttt{running})
            \ENDIF
        \ENDFOR
        
        \STATE \textit{\# Step 2: Handle swap In Requests}
        \IF{HasSwapIn(\texttt{swapped})}
            \STATE ExecuteSwapIn(\texttt{swapped})
            \STATE UpdateQueue(\texttt{r\_info}, \texttt{swapped}, ``SwapIn'')
        \ENDIF
        
        \STATE \textit{\# Step 3: Handle swap Out Requests}
        \IF{HasSwapOut(\texttt{running})}
            \STATE ExecuteSwapOut(\texttt{running})
            \STATE UpdateQueue(\texttt{r\_info}, \texttt{running}, ``SwapOut'')
            
            \STATE \textit{\# Step 3.1: Conflict Detection}
            \IF{DetectConflict(\texttt{running}, \texttt{swapped})}
                \STATE Synchronize(\texttt{running}, \texttt{swapped})
            \ENDIF
        \ENDIF
        \STATE \textit{\# Step 4: Dynamic Swapping (Async Or Sync)}
        \STATE decision $\gets$ \smash{Strategy(\texttt{running}, \texttt{swapped}, \texttt{r\_info})}
        \IF{decision == ``yes''}
            \STATE MovePending(\texttt{running}, \texttt{ongoing\_swap\_in})
        \ELSE
             \STATE SwapInStreamSynchronize()
        \ENDIF
    \ENDFOR
\end{algorithmic}
\end{algorithm}

\paragraph{Algorithm Explanation.}
\cref{alg:multi_threaded_swap_manager} details the operational workflow of the Multithreading Swap Manager. The process starts by monitoring ongoing swap in operations. Once an asynchronous swap in is completed, the corresponding request is moved from the \texttt{ongoing\_swap\_in} queue to the \texttt{running} queue, and the \texttt{r\_info} queue is updated accordingly. The manager then proceeds to handle both swap in and swap out requests. After completing swap out operations, it checks for any KV cache conflicts between the recently swap out requests and the ongoing swap in requests. If a conflict is detected, the manager performs synchronization to resolve the issue. Lastly, the algorithm employs a dynamic execution strategy, leveraging real-time profiling to optimize decision-making at each iteration. It evaluates recent swapping metrics from the \texttt{recent\_swap\_info} queue to determine whether to asynchronously proceed the swap in or to initiate a synchronization of the swap in stream to synchronize it.

\subsection{KV Cache Reuse Mechanism for Efficient Multi-turn Conversations}
\label{design:kv_cache_reuse}
In the last optimization of our design, to tackle \textbf{Challenge \#3}, we introduce the KV Cache Reuse Mechanism to reuse the KV cache copies in CPU and handle partial cache contamination, where the KV cache in CPU memory is contaminated by higher-priority requests. This mechanism enables the reuse of partially valid KV cache, minimizing preemption overhead by reducing the volume in resource-constrained scenarios.

\begin{figure}[ht]
    \centering
    \includegraphics[width=0.8\linewidth]{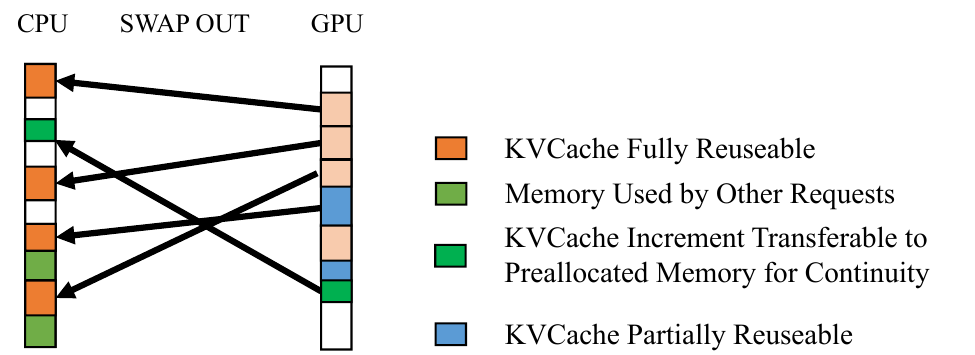}
    \caption{Workflow of the KV Cache Reuse Mechanism.}
    \label{fig:kv_cache_reuse_flow}
\end{figure}

\textbf{KV Cache Reuse Mechanism.}
Our mechanism focuses on retaining and efficiently managing KV cache by keeping a copy of the KV cache in CPU memory. To solve the \textbf{Challenge \#3}, we designed an algorithm that effectively tracks the status of KV cache. To ensure the validity of reused KV cache, we track and monitor which segments of the KV cache block group have been contaminated by higher-priority requests. So we can identified uncontaminated block groups that are eligible for reuse, preventing erroneous data access. As shown in \cref{fig:kv_cache_reuse_flow}, during preemption, the KV Cache Reuse Mechanism selectively swaps out only the necessary portions of the KV cache, minimizing the KV cache swapping between CPU and GPU memory and reducing preemption latency. Furthermore, the system preallocates additional memory space for the next turn's swap out increment, which is adjacent to KV cache copy already stored in CPU memory. This proactive allocation improves memory continuity, prevents fragmentation, and ensures smoother transitions between turns. In summary, compared to vLLM, this mechanism successfully reduces the swap out volume and preserves granularity by reusing partially valid KV cache and preallocation in constrained scenarios, leading to a significant reduction in preemption latency.

%% file: 4_methodology.tex
\section{Methodology}
\label{sec:methodology}

\textbf{System and Workload Configuration.} 
We evaluate \design{} using the LLaMA-8B and Qwen-32B models on NVIDIA A10 24 GB and A100 80 GB GPUs, respectively. Each GPU is configured with 60 GB of CPU swap space for KV cache offloading. The setup utilizes PCIe 4.0 with a x16 interface, enabling each GPU to achieve a theoretical bandwidth of 32 GB/s for data transfer to the CPU (64 GB/s bidirectional).

We utilize the Multi-Round ShareGPT~\cite{sharegpt} dataset to simulate extended, realistic conversations, as depicted in \cref{fig:sharegpt_distributions_cdf}. Since the output content is orthogonal to our work, we retain the original input and output lengths. From the dataset, we randomly select 1,000 multi-turn conversations with an average number of turns of 5.5 per conversation and generate request arrival traces based on a Poisson distribution with an average rate of 1 request per second.

\textbf{Context Switching Trace Simulation.} 
As there are no publicly available context-switching traces for Large Language Model as a Service (LLMaaS) workloads, we refer to the work of \cite{llmaas_arxiv2024} and simulate two patterns of context switching to evaluate system behavior under different conditions. These simulations include:
\textbf{Random.}
In this pattern, context switching occurs unpredictably, with adjustments of requests' priorities happening arbitrarily. There is no temporal correlation, simulating a dynamic and uncontrolled environment where the workload characteristics are highly unpredictable.
\textbf{Markov.}
This pattern introduces temporal locality into the context switching process. Requests that have been frequently or recently served are given higher priority, reflecting a more structured scenario where the workload exhibits some continuity or locality in its usage patterns.

In both cases, the priorities of requests are determined offline, meaning they are precomputed based on the simulation patterns rather than dynamically adjusted during runtime. The priority-update frequency determines how often the priorities are updated. When the frequency is set to 0.01, it means that every 100 iterations, the priorities of all requests are updated based on the current pattern. The scheduler then reorders requests across waiting, running, and swapped queues to meet the updated priority requirements. During other iterations, it adheres to the most recently updated priority to handle scheduling and service execution.

We evaluate the end-to-end performance of \design{} by comparing it with the baseline under appropriate priority-update frequency.
For Qwen-32B, we set the priority-update frequency to 0.02, following the study in Andes~\cite{andes_arxiv2024} to maximize the QoE for the round-robin pattern. On the other hand, for the smaller model LLaMA-8B, we double the priority-update frequency to 0.04 to better highlight the optimizations in context switching achieved by our design, especially under more frequent context switching.

\textbf{Baselines and Metrics.}
We compare \design{} with vLLM (0.3.3)~\cite{vllm_sosp23}. vLLM is a state-of-the-art LLM serving system optimized for throughput. The key metrics used for evaluation include the P95, P99, and P99.9 TTFT, which measure the latency experienced by the 95th, 99th, and 99.9th percentiles of requests before the first token of each turn is generated. In addition, we evaluate the P99.9 TBT, which captures the latency between consecutive tokens in the generated responses. Furthermore, we compare the end-to-end throughput of both systems. These metrics offer a comprehensive view of the system's performance, especially under different load conditions and priority schemes.

\textbf{Implementation.}
\design{} is built on vLLM with over 5,000 lines of Python and 1,000 lines of C++/CUDA code. On \design{} and vLLM, we first support multi-turn conversations utilizing ``prefill-with-prefix'' triton kernel from lightllm~\cite{lightllm_2024}. We then developed a priority-based scheduler that enhances existing scheduling policies by enabling dynamically adjusting priorities and managing request queues in real-time.

%% file: 5_evaluation.tex
\section{Evaluation}
\label{sec:evaluation}

\definecolor{baseline_2}{RGB}{127, 127, 255}  
\definecolor{feature_1}{RGB}{64, 200, 64}    
\definecolor{feature_2}{RGB}{255, 127, 127}   
\definecolor{feature_4}{RGB}{255, 165, 0}   

\begin{figure*}[ht]
    \centering
    \subfloat[LLaMA-8B \& Markov]{
        \centering
        \includegraphics[width=.32\linewidth, height=.24\linewidth]{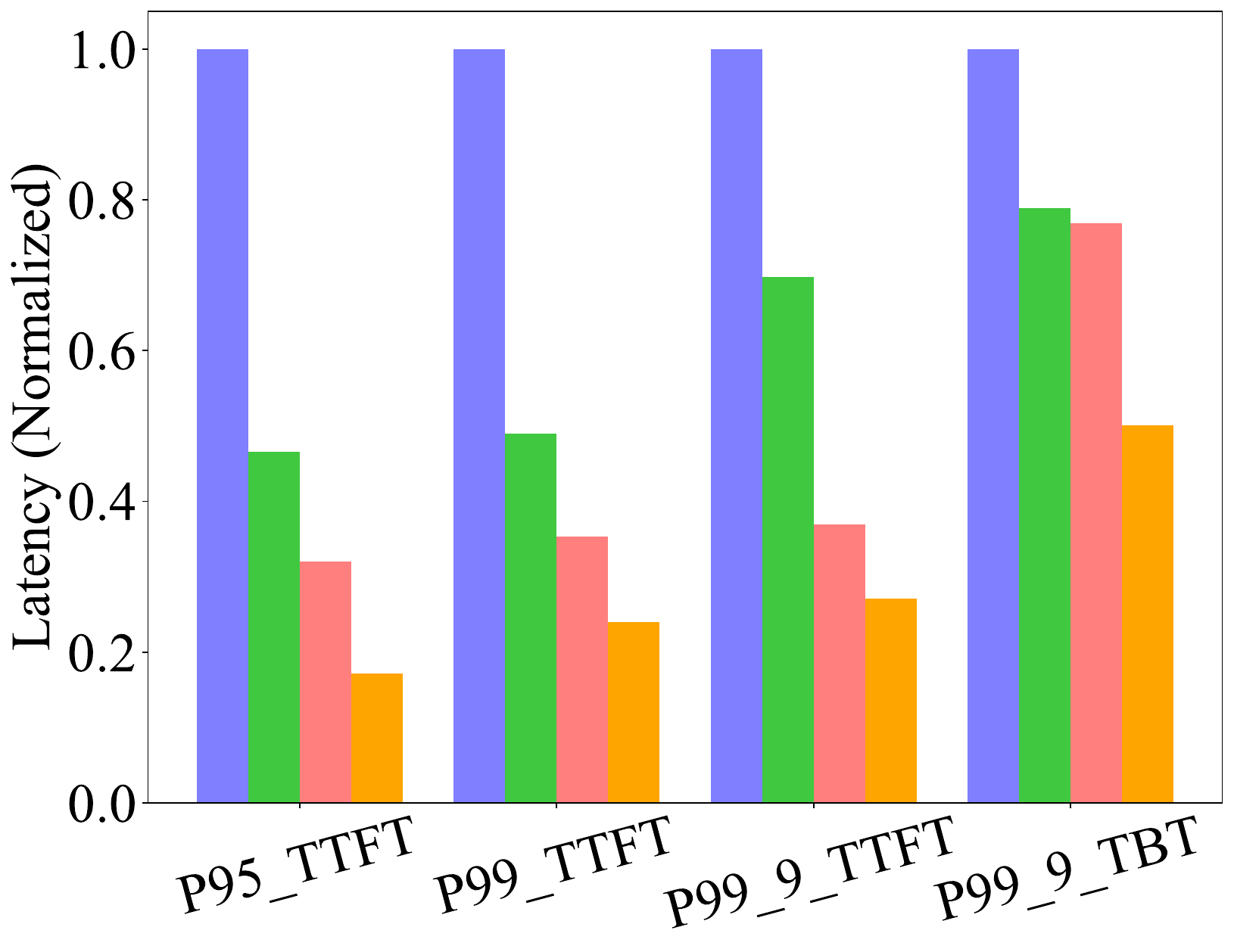}
    }
    \subfloat[LLaMA-8B \& Random]{
        \centering
        \includegraphics[width=.32\linewidth, height=.24\linewidth]{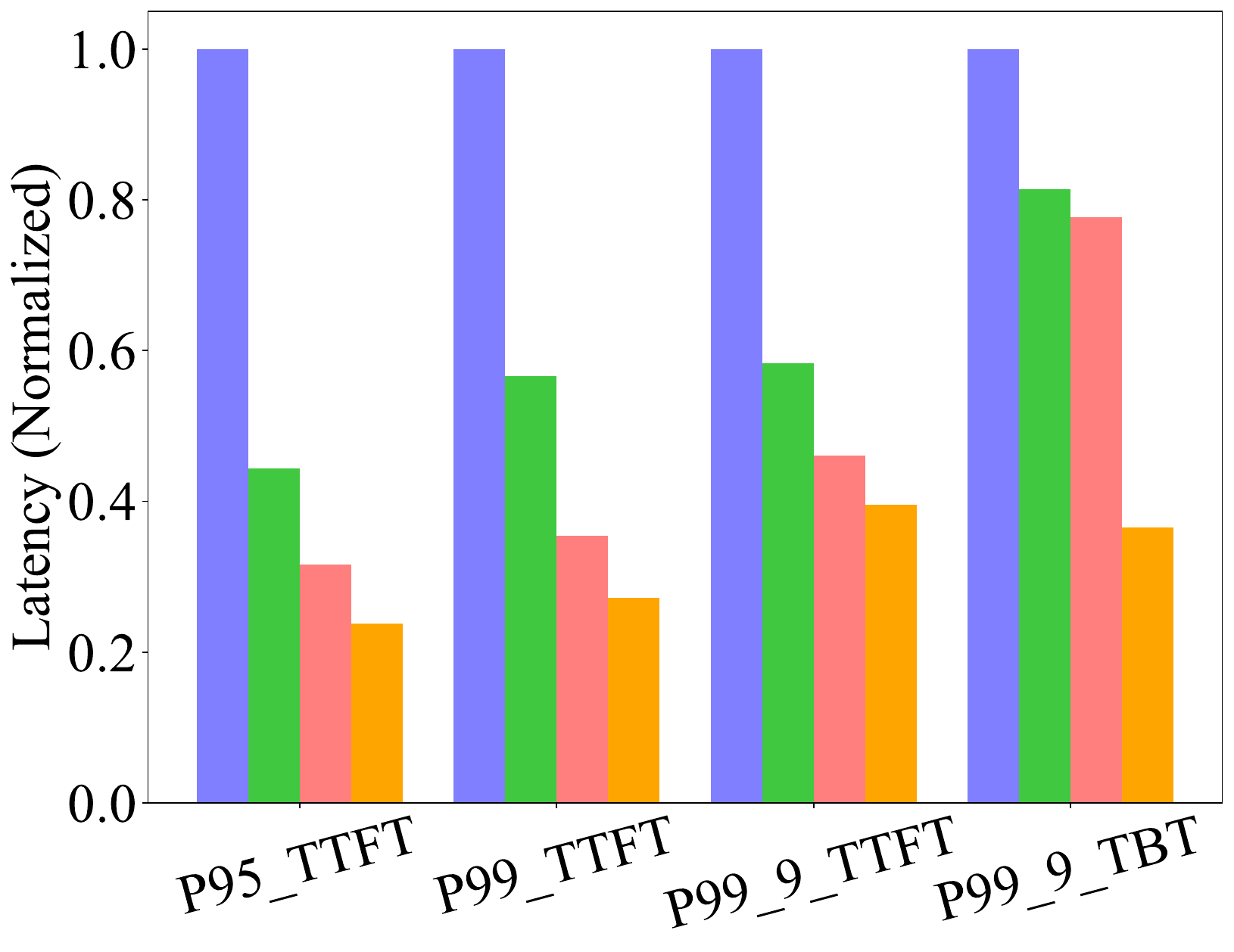}
    }
    \subfloat[Qwen-32B \& Markov]{
        \centering
        \includegraphics[width=.32\linewidth, height=.24\linewidth]{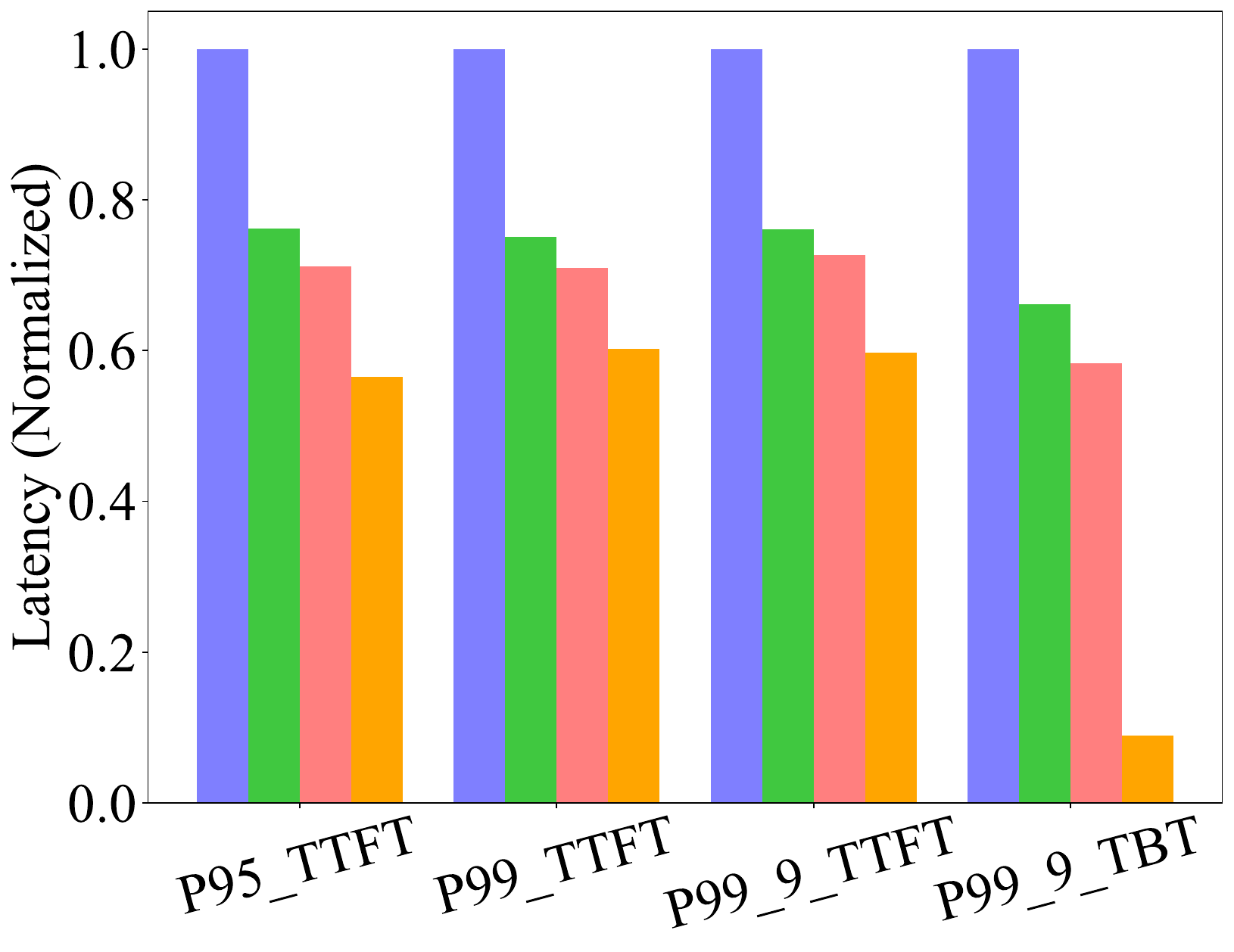}
    }

    \subfloat[Qwen-32B \& Random]{
        \centering
        \includegraphics[width=.32\linewidth, height=.24\linewidth]{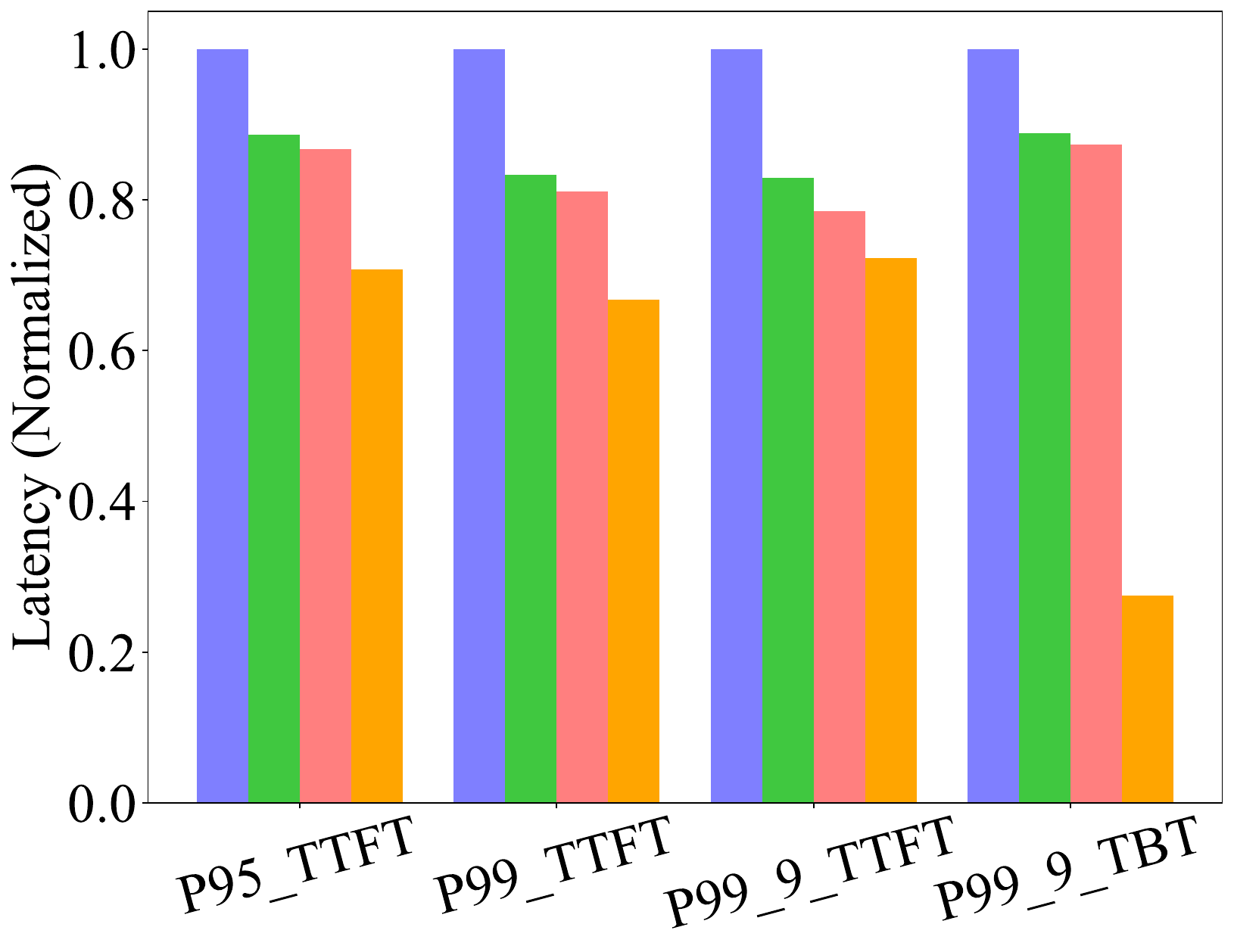}
    }
    \subfloat[LLaMA-8B Throughput]{
        \centering
        \includegraphics[width=.32\linewidth, height=.24\linewidth]{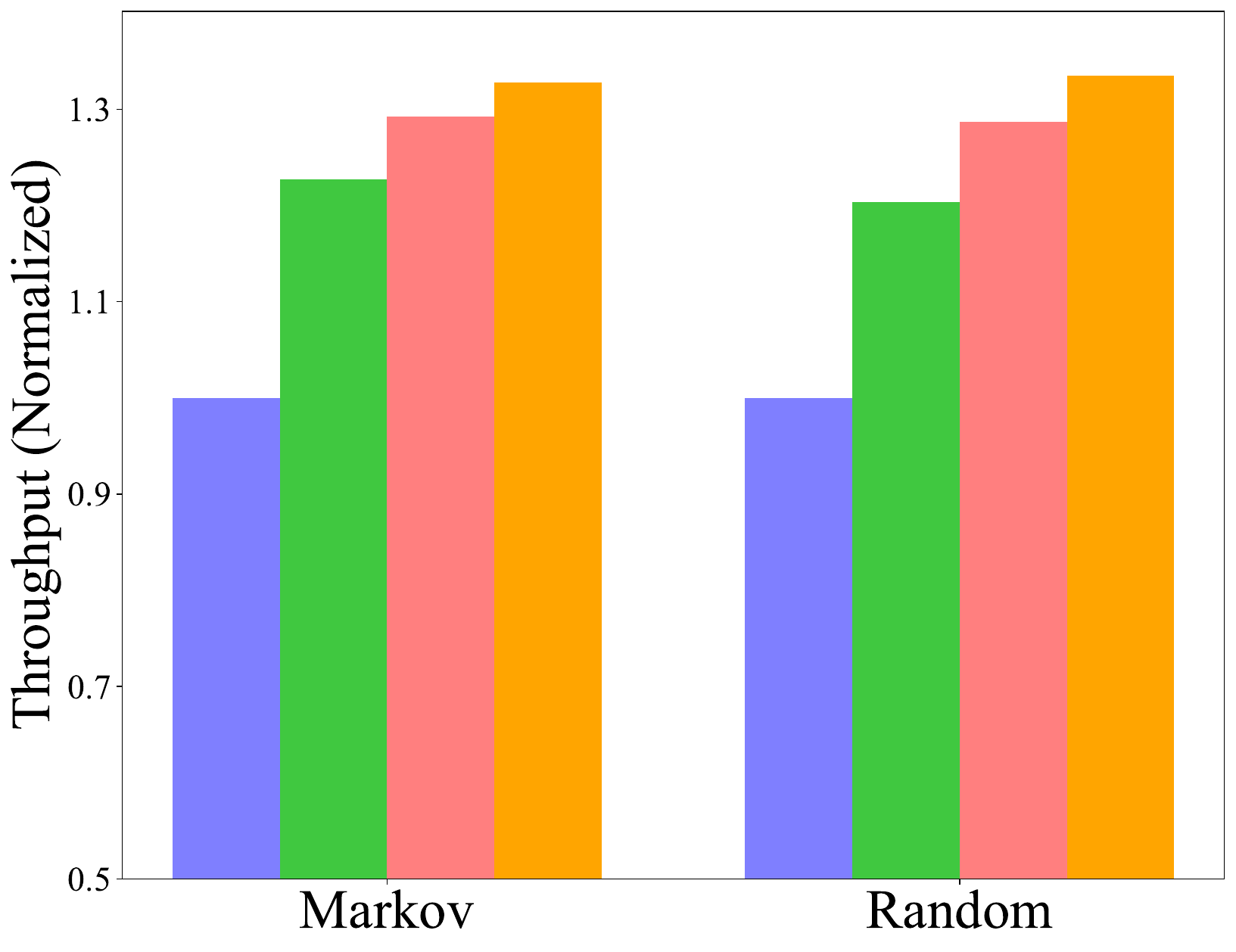}
    }
    \subfloat[Qwen-32B Throughput]{
        \centering
        \includegraphics[width=.32\linewidth, height=.24\linewidth]{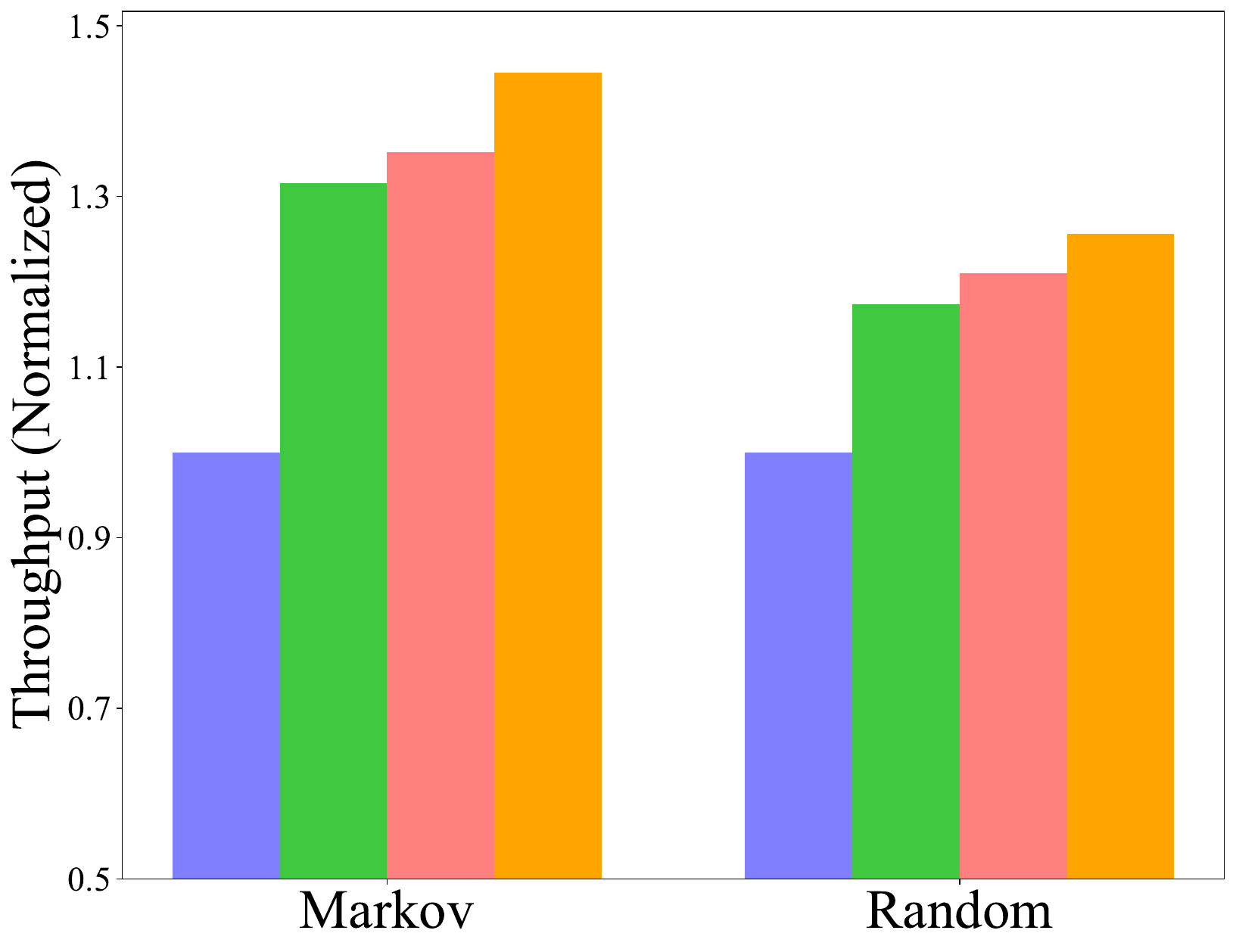}
    }
    \vskip\baselineskip
    \centering
    \begin{tikzpicture}
        \draw[baseline_2, fill=baseline_2] (0,0) rectangle (0.3,0.3);
        \node[anchor=west] at (0.30,0.15) {vLLM};

        \draw[feature_1, fill=feature_1] (1.75,0) rectangle (2.05,0.3);
        \node[anchor=west] at (2.1,0.15) {Dynamic Block Group Only};

        \draw[feature_2, fill=feature_2] (6.75,0) rectangle (7.05,0.3);
        \node[anchor=west] at (7.10,0.15) {\makecell{Dynamic Block Group \\ + KV Cache Reuse}};

        \draw[feature_4, fill=feature_4] (11,0) rectangle (11.3,0.3);
        \node[anchor=west] at (11.30,0.15) {\design{}};
    \end{tikzpicture}
    \centering\caption{Comparison of TTFT, TBT, and throughput between \design{} and baseline under different models and traces.}
    \label{fig:end2end_experiment_results}
\end{figure*}
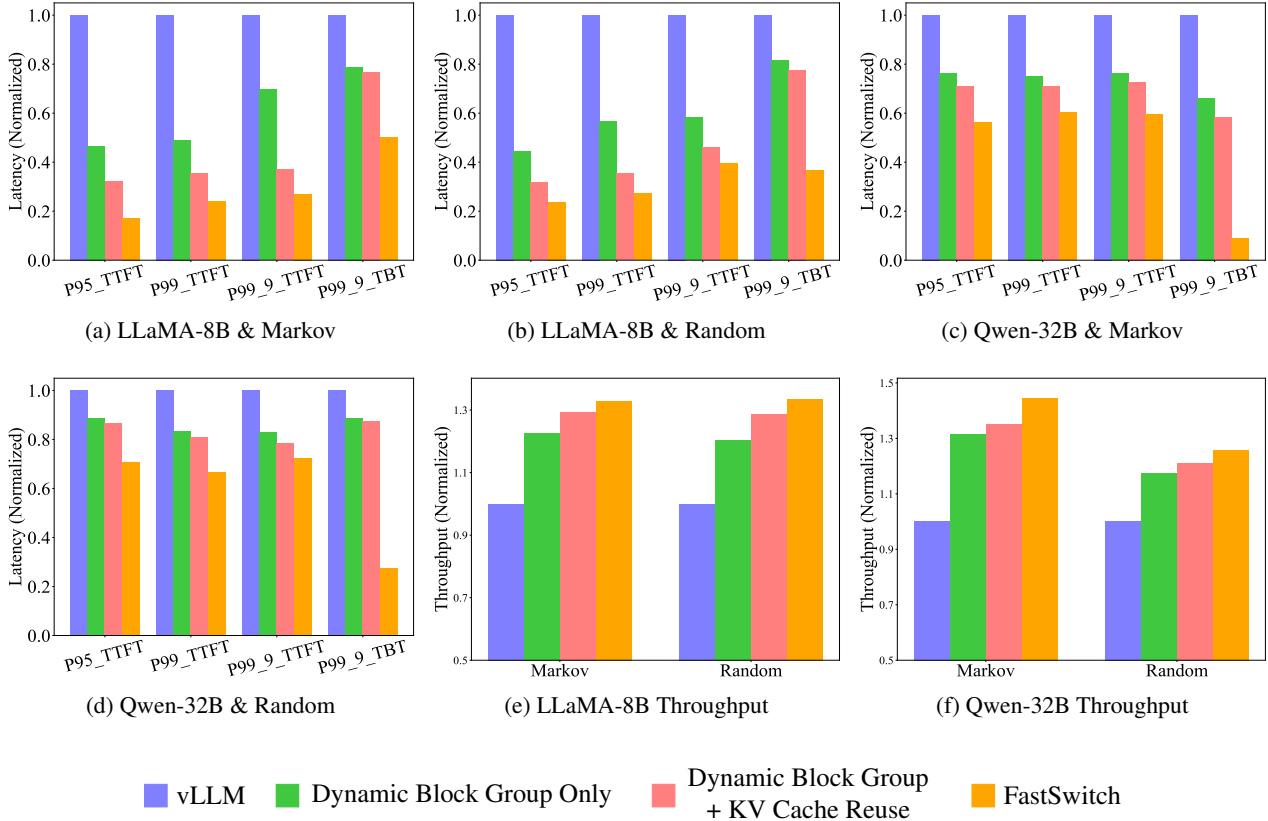

\subsection{End-to-End Performance}
\label{subsec:end-to-end performance}

\subsubsection{Latency Metrics}

We start by analyzing SLO metrics for both the LLaMA-8B and Qwen-32B models under Markov and Random context-switching patterns. These metrics include P95, P99, P99.9 TTFT, and P99.9 TBT.
As shown in \cref{fig:end2end_experiment_results} (a)--(d),  
overall, under the high priority-update frequency, \design{} demonstrates significantly lower latency compared to other designs. This indicates that \design{} can effectively leverage priority adjustments to meet the SLOs of more users simultaneously without incurring significant overhead.
Specifically, across different pattern,\design{} achieves speedups of 4.3-5.8$\times$, 3.7-4.1$\times$, 2.5-3.7$\times$, and 2.0-2.7$\times$ in P95 TTFT, P99 TTFT, P99.9 TTFT, and P99.9 TBT, respectively, for LLaMA-8b. And for Qwen-32B, it achieves speedups of 1.4-1.7$\times$, 1.5-1.6$\times$, 1.3-1.4$\times$, and 3.6-11.2$\times$ for P95 TTFT, P99 TTFT, P99.9 TTFT, and P99.9 TBT, respectively. 
To understand the impact of proposed optimizations for \design{}, we evaluate them incrementally on top of the baseline vLLM. The figure shows the latency normalized to vLLM for each optimization by adding it on top of the former one. Our observations in Qwen-32B show that using Dynamic Block Group achieves an average of 1.3× improvement across different SLOs compared to vLLM due to significant improvement in I/O utilization; moreover, adding the KV Cache Reuse Mechanism to the Dynamic Block Group Manager further improves various SLOs latency, indicating that the KV Cache Reuse Mechanism effectively eliminates redundant I/O transfers; furthermore, by incorporating the Multithreading Swap Manager, \design{} achieves an average 1.42× improvement across various SLOs, demonstrating the effectiveness of this optimization in improving GPU utilization.

It is worth noting that, overall, prefill tends to take longer than decode, meaning that the same swapping delay before inference has a greater impact on TBT than TTFT. This effect becomes increasingly pronounced as the model size increases, due to the growing ratio of prefill to decode overhead. Under the Random pattern, swapping becomes more intense compared to the Markov one. This is because the Markov pattern tends to retain more recent requests within the running batch, whereas the Random pattern does not, disrupting the continuity of the block group. Additionally, it increases KV cache conflicts that results in more synchronization and reduces the reuse of KV cache in the CPU. Despite these problems, our approach still achieves significant performance gains. Overall, with longer contexts and larger models, the inference time—due to its memory-bound nature—does not grow as quickly as the overhead caused by swapping. This amplifies the overhead caused by context switching, making it a more significant bottleneck. Consequently, our I/O optimizations become increasingly impactful in improving overall SLOs in these cases.

\subsubsection{End-to-End Throughput Improvement}

In addition to latency metrics, we evaluate the end-to-end throughput of \design{} under varying priority-update frequencies for both models. As illustrated in \cref{fig:end2end_experiment_results} (e)--(f), \design{} consistently enhances throughput across different priority-update frequencies.

For the LLaMA-8B model, \design{} achieves up to a 1.334$\times$ increase in throughput under high priority-update frequency across both patterns, maintaining efficient token generation without significant delays. Similarly, the Qwen-32B model experiences up to a 1.444$\times$ improvement in throughput. The larger throughput gains observed for Qwen-32B are attributed to its higher swapping latency compared to its inference time, which \design{} effectively mitigates through the three optimizations.

\subsubsection{Summary}

The contributions of each sub-design in FastSwitch, are critical in reducing latency and improving throughput under frequent priority adjustments. \design{} consistently outperforms other designs across different models and context switching patterns, demonstrating its scalability and effectiveness in managing complex workloads.

\begin{figure}[ht]
    \centering
    \includegraphics[width=0.9\linewidth]{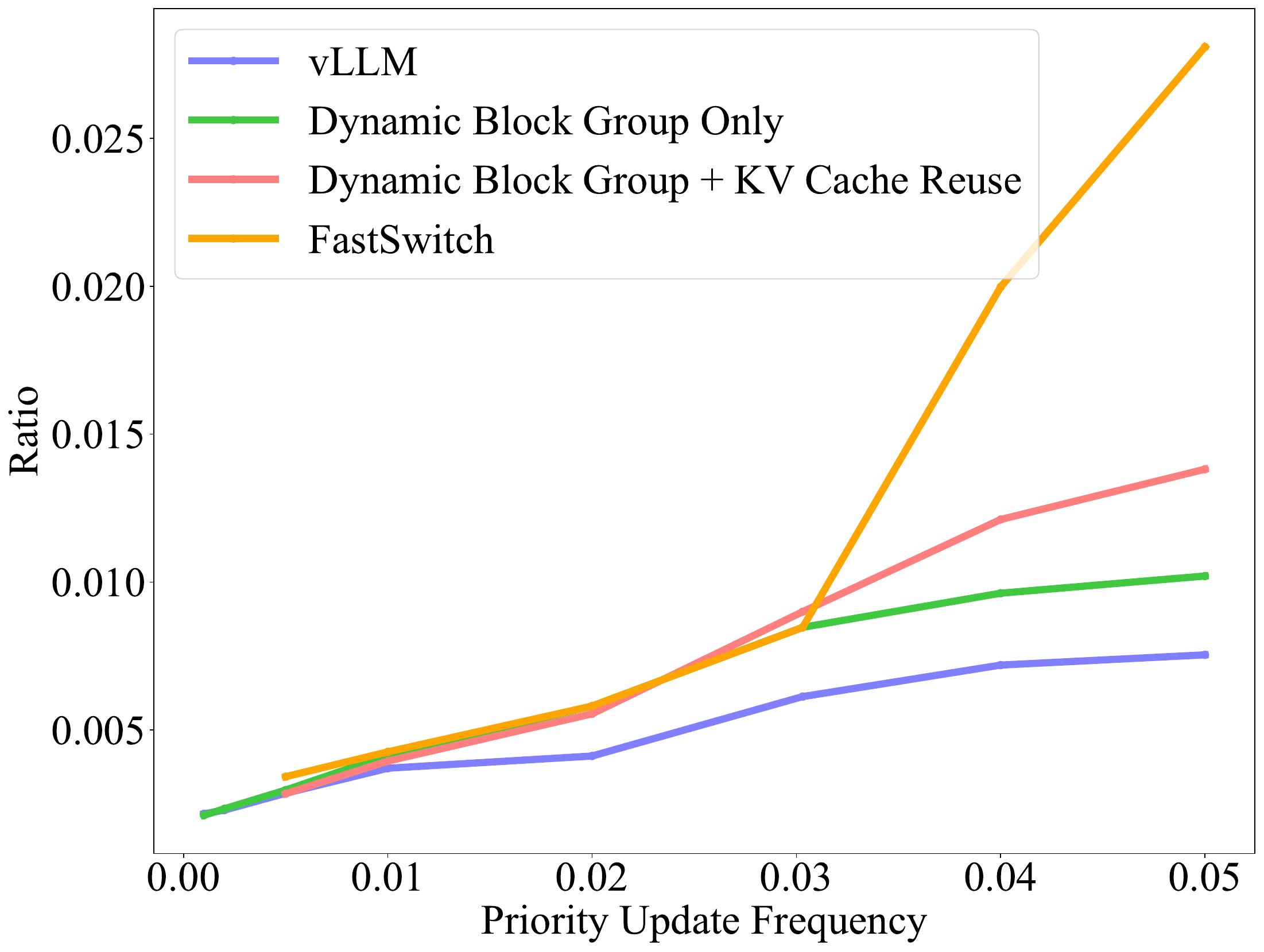}
    \caption{Show the call stack overhead after applying each optimization.}
    \label{fig:call_stack_overhead}
\end{figure}

\subsection{Call Stack Overhead Analysis}
\label{subsec:call_stack_overhead}

\cref{fig:call_stack_overhead} illustrates the call stack overhead as priority-update frequency increases. Each part of \design{} introduces optimizations that slightly raise overhead but improve performance. Despite the gradual increase, the overhead contributes to no more than a 1 \% rise in end-to-end time. This indicates a minimal impact on overall system efficiency. As the frequency increases, the call stack overhead also increases. This is due to the need to resolve more KV cache conflicts and synchronize more ongoing swap-in requests dispatched in pass iterations by the end of the schedule. This results in some context switching overhead being added to the call stack overhead. However, pure call stack overhead remains within 1 \%.

\subsection{Breakdown and Sensitivity Analysis}
\label{subsec:breakdown_analysis}

To gain deeper insights into the effectiveness of our proposed system and its sensitivity to various parameters, we perform a series of breakdown analysis experiments using the LLaMA-8B model on the ShareGPT dataset, running on a single A10 GPU. The average request rate is set to 1.0 req/s.

\subsubsection{Dynamic Block Group Manager}
\label{subsubsec:dynamic_block_group_manager_evaluation}

\begin{figure}[htbp]
    \centering
    \begin{minipage}[t]{0.485\linewidth}
        \centering
        \includegraphics[width=\linewidth]{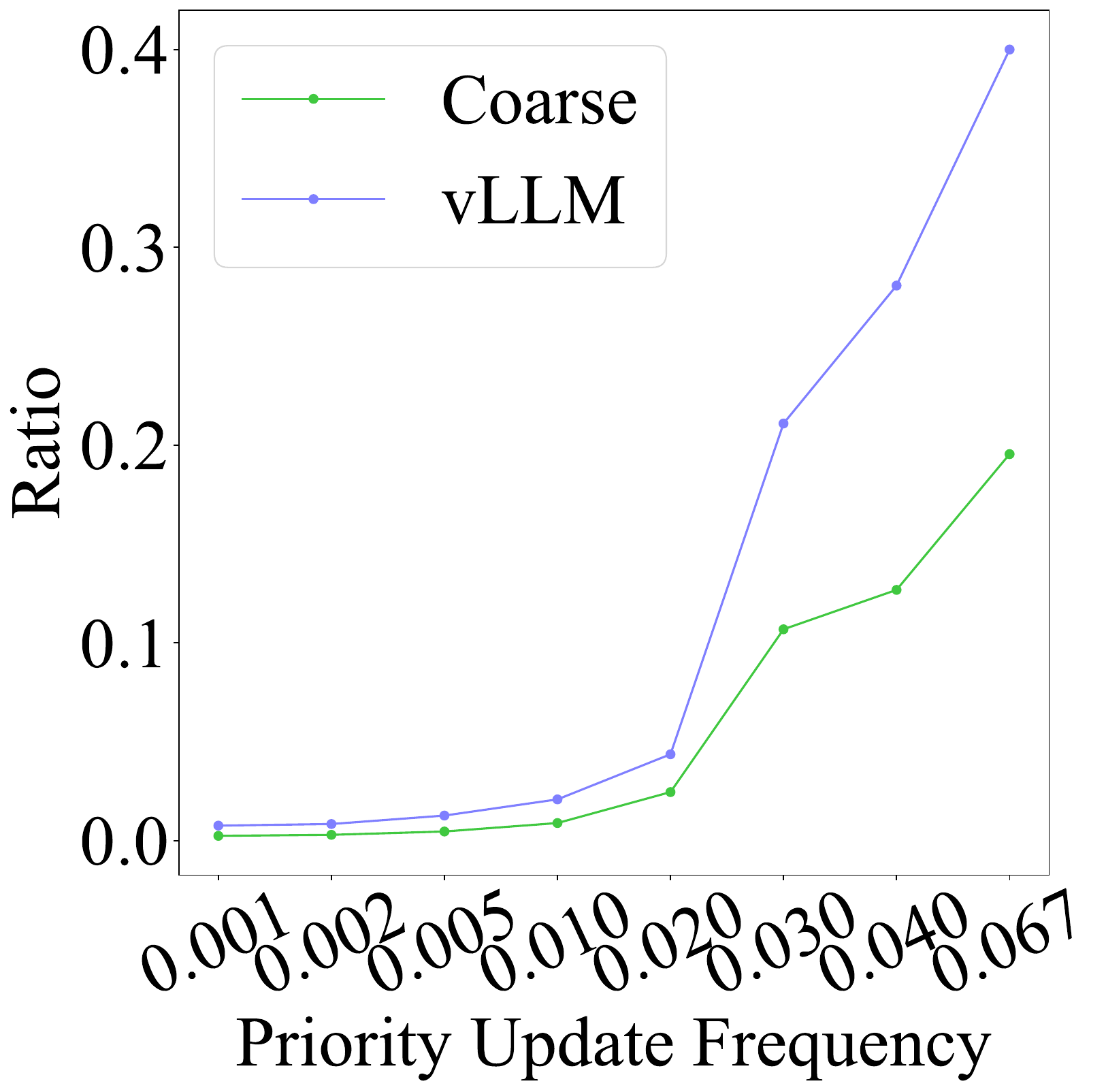}
        \caption{Context switching overhead across priority-update frequencies.}
        \label{fig:context_switch_overhead_comparison}
    \end{minipage}
    \hspace{0.01\linewidth} 
    \begin{minipage}[t]{0.485\linewidth}
        \centering
        \includegraphics[width=0.98\linewidth]{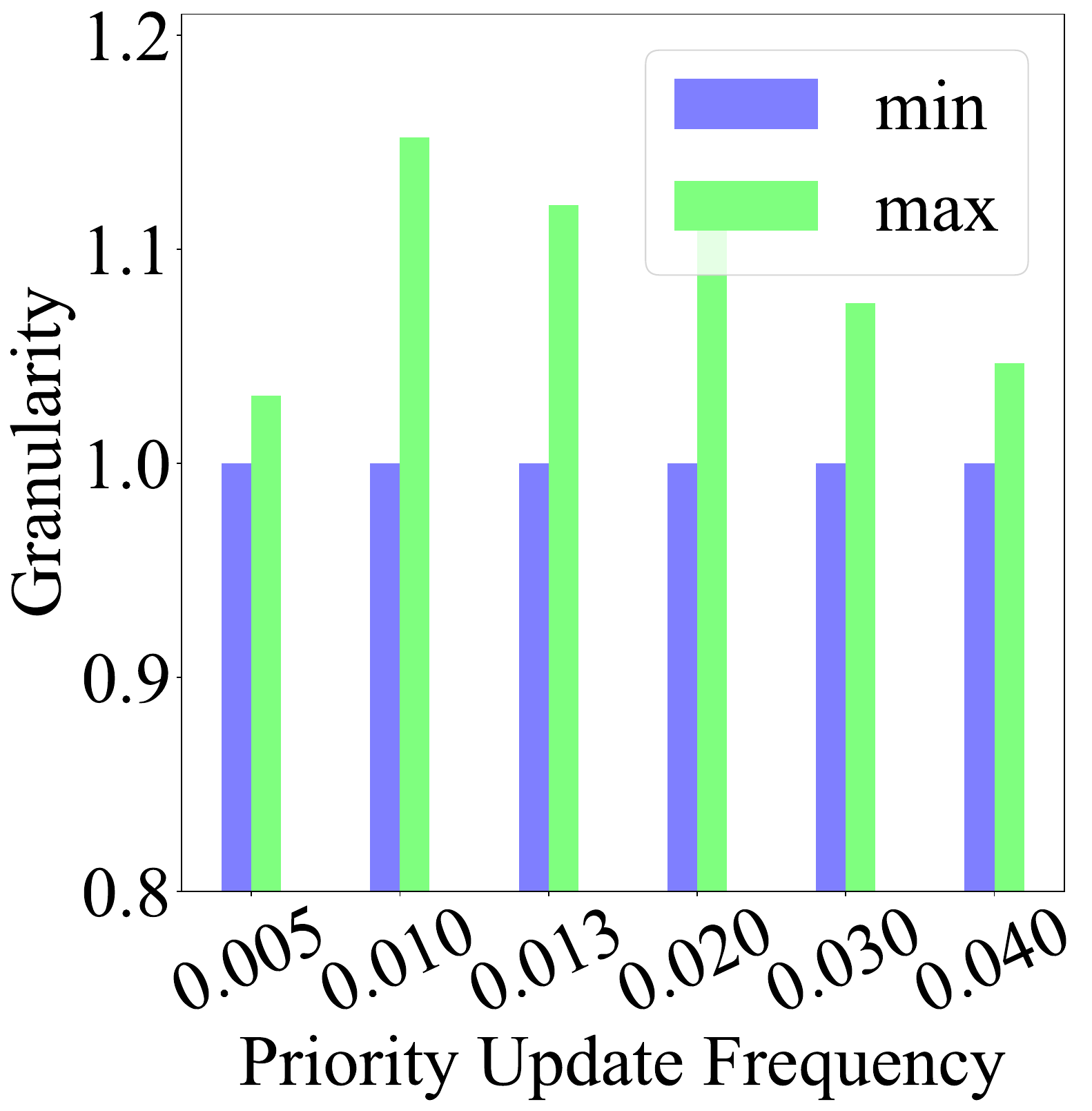}
        \caption{Sensitivity study.}
        \label{fig:dynamic_block_group_manager_initial_block_group_size_sensitivity_study}
    \end{minipage}
\end{figure}

\paragraph{Effectiveness of the Dynamic Block Group Manager.}
The Dynamic Block Group Manager employs a coarse-grained KV cache allocation approach, resulting in simplified swapping operations and reduced context switching overhead, as demonstrated by the ratio of context switching overhead to end-to-end latency shown in \cref{fig:context_switch_overhead_comparison}. The coarse-grained approach shows up to $3.11\times$ context switching speedup compared to the vLLM baseline across various frequencies.

\paragraph{Initial Dynamic Block Group Size Sensitivity.}
We set the initial block group size to 1,000 tokens, or about 70 vLLM blocks. A sensitivity analysis, shown in \cref{fig:dynamic_block_group_manager_initial_block_group_size_sensitivity_study}, examines the average swap in and swap out granularity across initial block group sizes (64 to 3,000 tokens) and varying priority update frequencies. The values are normalized, with the minimum set to 1. The results show that for a fixed priority-update frequency, changing the initial block group size causes no more than a 15.13\% difference in granularity. This indicates that the system is robust to variations in block group size. 
Granularity is mainly influenced by the GPU memory allocated to the KV cache for each task, making GPU memory a key factor in swapping efficiency.

\begin{figure}[htbp]
    \centering
    \begin{minipage}[t]{0.485\linewidth}
        \centering
        \includegraphics[width=0.937\linewidth]{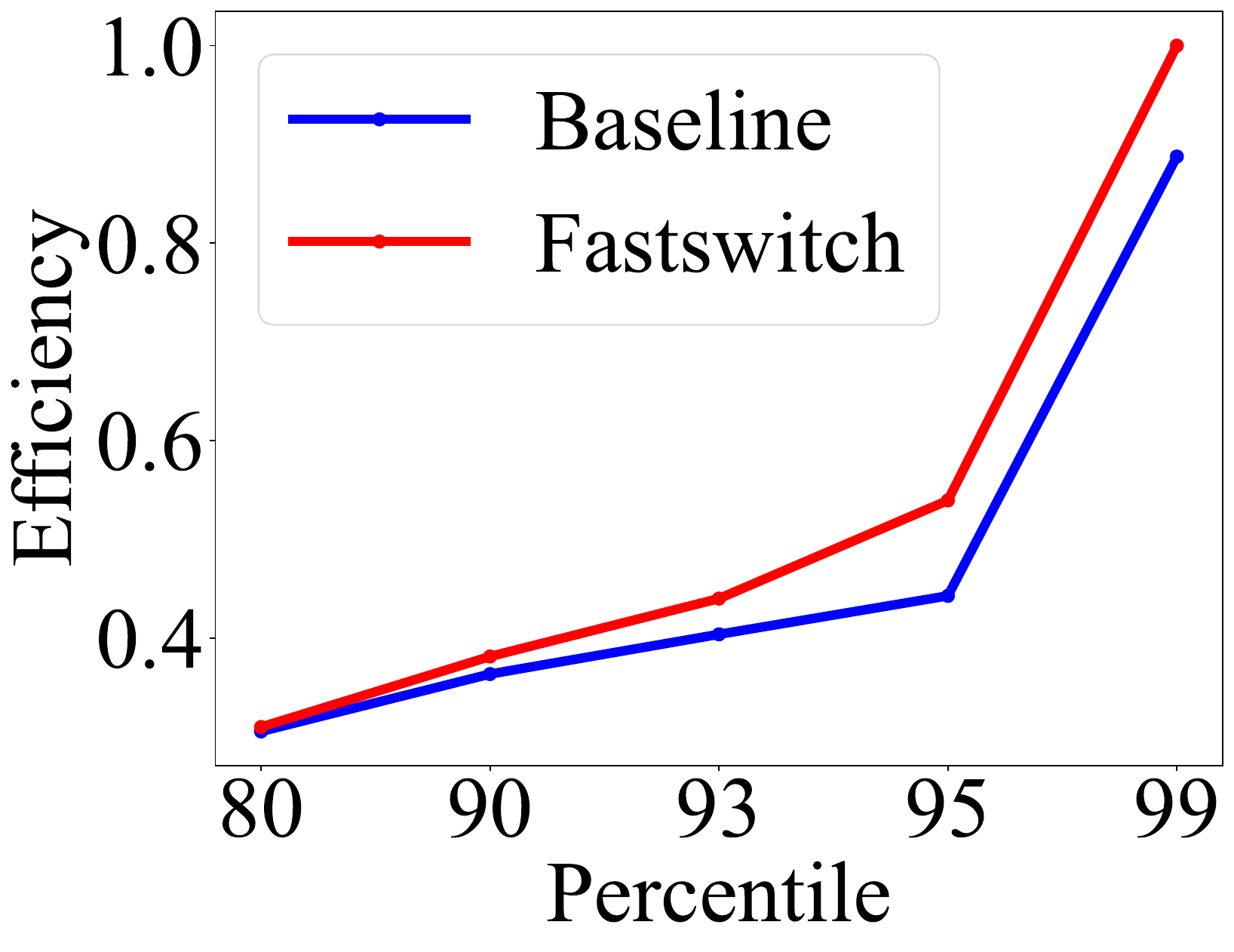}
        \caption{Efficiency comparison.}
        \label{fig:method2_token_generation}
    \end{minipage}
    \hspace{0.01\linewidth} 
    \begin{minipage}[t]{0.485\linewidth}
        \centering
        \includegraphics[width=\linewidth]{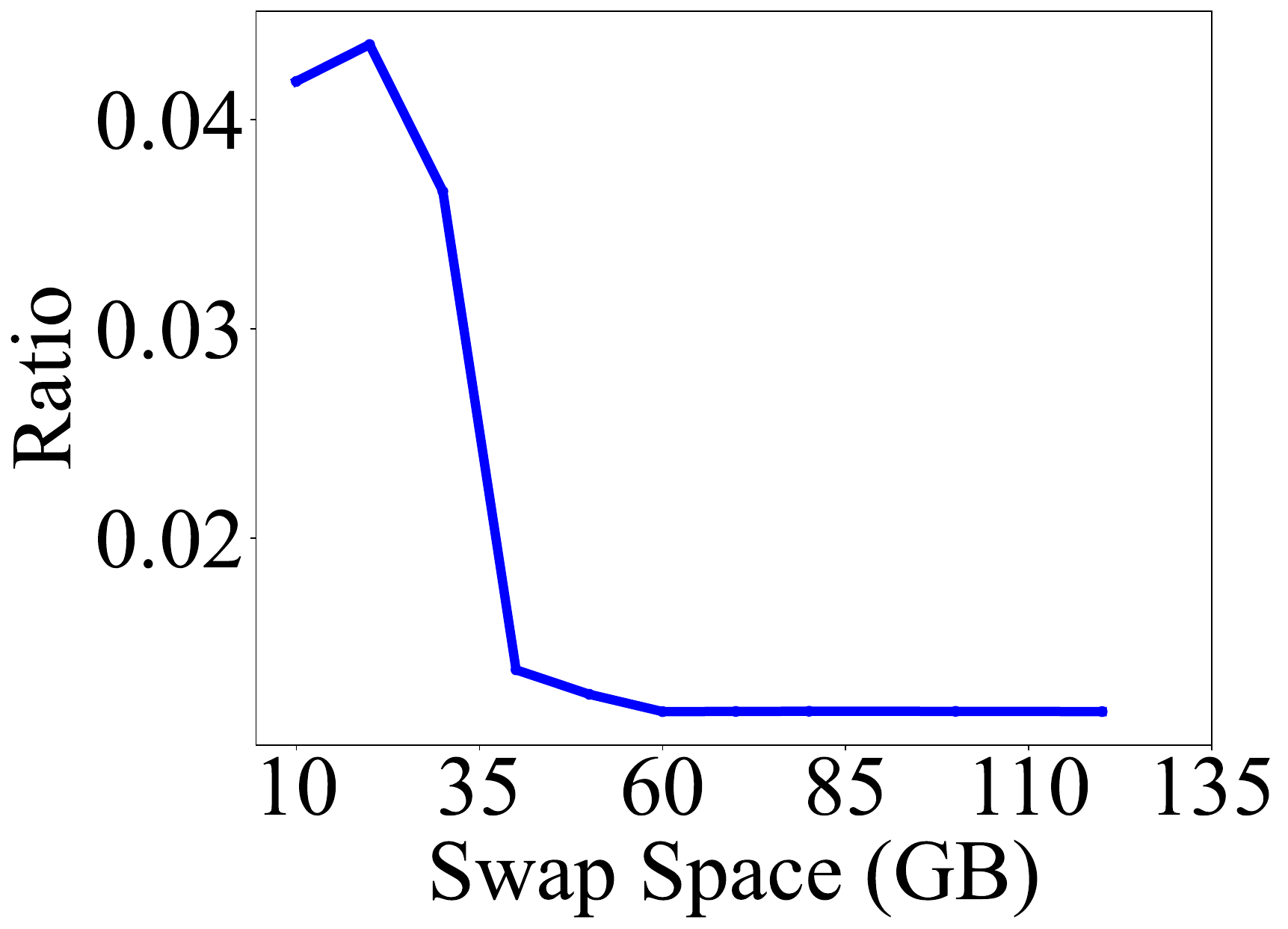}
        \caption{Sensitivity study.}
        \label{fig:cpu_memory_sensitivity}
    \end{minipage}
\end{figure}

\subsubsection{Multithreading Swap Manager}
\label{subsubsec:method2_evaluation}

\paragraph{Token Generation Efficiency.}
We define Token Generation Efficiency as the number of tokens generated per unit time. The introduction of the Multithreading Swap Manager, which facilitates asynchronous swapping, enhances this efficiency because of less GPU idle time. To compare the token generation efficiency between the baseline system (which includes other optimizations but does not use the swap manager) and our proposed design (\design{}), we conducted an analysis, as presented in \cref{fig:method2_token_generation}. However, the increased number of iterations introduced by the Multithreading Swap Manager complicates a direct comparison with the baseline. To address this, we divided the inference process into fixed-iteration intervals, each consisting of 5 iterations. Within each interval, we calculated the number of new tokens generated and the time taken, allowing us to determine the token generation efficiency per unit time for each interval. We then compared the token generation efficiency across different percentiles for both the baseline and \design{}. The results show that \design{} consistently achieves higher token generation efficiency across almost all percentiles, with particularly significant improvements at higher percentiles - showing a 21.8\% increase at P99 and a 12.6\% increase at P99.9 compared to the baseline. These results demonstrate the effectiveness of the Multithreading Swap Manager in minimizing the impact of KV cache transfers with frequent priority adjustments.

\begin{table}[ht]
    \centering
    \caption{Microbenchmark comparison.}
    \scriptsize  
    \setlength{\tabcolsep}{5pt} 
    \renewcommand{\arraystretch}{1} 
    \begin{tabular}{p{1.65cm} c c}
        \toprule
        \textbf{Metric} & \textbf{Traditional Swap Out} & \textbf{\makecell{Optimized Swap Out with \\ KV Cache Reuse}} \\
        \midrule
        Num blocks & 122030 & 58187 \\ 
        Num operations & 13076 & 10713 \\ 
        Latency & 15.5s & 6.7s \\ 
        \bottomrule
    \end{tabular}
    \label{tab:reduction}
\end{table}

\subsubsection{KV Cache Reuse Mechanism}
\label{subsubsec:method3_evaluation}

\paragraph{Reduction in Swap Out Volume.}
The KV Cache Reuse Mechanism significantly reduces the total number of swap out blocks by 53\%, as shown in \cref{tab:reduction}, directly correlating with lower preemption latency and stalled time. 

\paragraph{Sensitivity of CPU Memory Size on Context Switching Overhead.}
We evaluated how varying CPU memory sizes for KV cache copies impact the KV Cache Reuse Mechanism by measuring context switching overhead. As shown in \cref{fig:cpu_memory_sensitivity}, increasing memory allows less KV cache copies to be contaminated, reducing context switching overhead by enabling greater cache reuse and minimizing redundant KV cache swapping

Our findings show that larger memory allocation reduce overhead by supporting more cache reuse across conversation turns. However, beyond 60 GB, further increases yield diminishing returns, suggesting 60 GB as an optimal allocation for this setup.

%% file: 6_related_work.tex
\section{Related Work}
\subsection{Scheduling, SLOs, and Fairness in LLM Serving}

Maintaining fairness and meeting SLOs in LLM serving systems is crucial to ensuring overall service quality. Sheng et al.~\cite{fairness_osdi24} proposed VTC, a fair scheduler for LLM serving that handles unpredictable request lengths and dynamic batching, ensuring fairness at the token level. Andes~\cite{andes_arxiv2024} introduces Quality-of-Experience (QOE) which balances latency and quality in LLM-based text streaming services through the adjustments of requests' priorities. FastServe~\cite{fastserve_arxiv23} introduced a fast distribute inference serving system that improves scheduling and resource management in distributed environments. However, priority adjustments made to achieve better SLOs may introduce context switching overhead, which can negatively impact SLOs and offset the benefits of priority adjustments.

\subsection{KV Cache Management}
KV cache~\cite{cachegen_sigcomm24, mooncake_arxiv2024, model_arxiv2023} stores precomputed key-value projections from previous tokens during Transformer model inference. Efficient KV cache management improves the throughput in LLM serving system, particularly batched executions and multi-turn conversation. The cache holds intermediate key-value pairs for self-attention~\cite{self_arxiv2018}, addressing the issue of redundant computation during token generation.
vLLM~\cite{vllm_sosp23} employs paging for memory-efficient KV cache management, enabling large batch processing through dynamic GPU memory paging. SGLang~\cite{sglang_arxiv2023} introduces RadixAttention, organizing KV cache in a radix tree for systematic cache reuse across shared-prefix requests, with LRU-based eviction. Unlike vLLM's batch-level focus, RadixAttention handles both intra-program parallelism and multi-call workflows.
Other systems like TensorRT-LLM~\cite{tensorrt_llm} and Hugging Face Accelerate~\cite{huggingface_llm} optimize throughput via dynamic batch sizing, but lack fine-grained prefix reuse and intra-program parallelism capabilities found in vLLM and SGLang.
However, these systems overlook the challenges introduced by page memory-based memory management and scheduling strategies for KV cache transmission in fairness-aware scenarios with frequent priority adjustments. To the best of our knowledge, \design{} is the first system to focus on the significant overhead of context switching and propose comprehensive optimizations to address it.

%% file: 7_conclusion.tex
\section{Conclusion}
We present \design{}, a serving system that optimizes I/O usage and GPU utilization through our KV cache management and scheduling to ensure fairness without sacrifising SLOs. On top of \design{}, this work proposes optimizations focusing on addressing three identified challenges, poor I/O utilization, GPU idleness, and redundant I/O transmission.
Our evaluation shows that \design{} outperforms the state-of-the-art LLM serving system vLLM with speedups of 1.4-11.2$\times$ across various tail TTFT and TBT.

%% file: main.bbl
\begin{thebibliography}{37}
\providecommand{\natexlab}[1]{#1}
\providecommand{\url}[1]{\texttt{#1}}
\expandafter\ifx\csname urlstyle\endcsname\relax
  \providecommand{\doi}[1]{doi: #1}\else
  \providecommand{\doi}{doi: \begingroup \urlstyle{rm}\Url}\fi

\bibitem[Agrawal et~al.(2023)Agrawal, Panwar, Mohan, Kwatra, Gulavani, and Ramjee]{sarathi_arxiv2023}
Agrawal, A., Panwar, A., Mohan, J., Kwatra, N., Gulavani, B.~S., and Ramjee, R.
\newblock Sarathi: Efficient llm inference by piggybacking decodes with chunked prefills.
\newblock \emph{arXiv preprint arXiv:2308.16369}, 2023.

\bibitem[Agrawal et~al.(2024)Agrawal, Kedia, Panwar, Mohan, Kwatra, Gulavani, Tumanov, and Ramjee]{agrawal_arxiv2023}
Agrawal, A., Kedia, N., Panwar, A., Mohan, J., Kwatra, N., Gulavani, B.~S., Tumanov, A., and Ramjee, R.
\newblock Taming throughput-latency tradeoff in llm inference with sarathi-serve.
\newblock \emph{arXiv preprint arXiv:2403.02310}, 2024.

\bibitem[Aminabadi et~al.(2022)Aminabadi, Rajbhandari, Zhang, Awan, Li, Li, Zheng, Rasley, Smith, Ruwase, and He]{deepspeed_arxiv2022}
Aminabadi, R.~Y., Rajbhandari, S., Zhang, M., Awan, A.~A., Li, C., Li, D., Zheng, E., Rasley, J., Smith, S., Ruwase, O., and He, Y.
\newblock Deepspeed inference: Enabling efficient inference of transformer models at unprecedented scale, 2022.
\newblock URL \url{https://arxiv.org/abs/2207.00032}.

\bibitem[Bisk et~al.(2020)Bisk, Zellers, Gao, Choi, et~al.]{piqa_aaai20}
Bisk, Y., Zellers, R., Gao, J., Choi, Y., et~al.
\newblock Piqa: Reasoning about physical commonsense in natural language.
\newblock In \emph{Proceedings of the AAAI conference on artificial intelligence}, volume~34, pp.\  7432--7439, 2020.

\bibitem[Brown(2020)]{chatgp3_arxiv2023}
Brown, T.~B.
\newblock Language models are few-shot learners.
\newblock \emph{arXiv preprint arXiv:2005.14165}, 2020.

\bibitem[Gao et~al.(2024)Gao, He, Sharma, Kang, Jevdjic, Deng, Yang, Yu, and Zuo]{attentionstore_arxiv2024}
Gao, B., He, Z., Sharma, P., Kang, Q., Jevdjic, D., Deng, J., Yang, X., Yu, Z., and Zuo, P.
\newblock Attentionstore: Cost-effective attention reuse across multi-turn conversations in large language model serving.
\newblock \emph{arXiv preprint arXiv:2403.19708}, 2024.

\bibitem[Ge et~al.(2023)Ge, Zhang, Liu, Zhang, Han, and Gao]{model_arxiv2023}
Ge, S., Zhang, Y., Liu, L., Zhang, M., Han, J., and Gao, J.
\newblock Model tells you what to discard: Adaptive kv cache compression for llms.
\newblock \emph{arXiv preprint arXiv:2310.01801}, 2023.

\bibitem[Gong et~al.(2018)Gong, He, Tan, Qin, Wang, and Liu]{frage_arxiv2018}
Gong, C., He, D., Tan, X., Qin, T., Wang, L., and Liu, T.-Y.
\newblock Frage: Frequency-agnostic word representation.
\newblock \emph{Advances in neural information processing systems}, 31, 2018.

\bibitem[Hendrycks et~al.(2020)Hendrycks, Burns, Basart, Zou, Mazeika, Song, and Steinhardt]{measuring_arxiv2020}
Hendrycks, D., Burns, C., Basart, S., Zou, A., Mazeika, M., Song, D., and Steinhardt, J.
\newblock Measuring massive multitask language understanding.
\newblock \emph{arXiv preprint arXiv:2009.03300}, 2020.

\bibitem[{Hugging Face}(2024)]{huggingface_llm}
{Hugging Face}.
\newblock Hugging face large language models (llms).
\newblock \url{https://huggingface.co/}, 2024.
\newblock Accessed: 2024-10-28.

\bibitem[Koh et~al.(2024)Koh, Fried, and Salakhutdinov]{multi_nips24}
Koh, J.~Y., Fried, D., and Salakhutdinov, R.~R.
\newblock Generating images with multimodal language models.
\newblock \emph{Advances in Neural Information Processing Systems}, 36, 2024.

\bibitem[Kwon et~al.(2023)Kwon, Li, Zhuang, Sheng, Zheng, Yu, Gonzalez, Zhang, and Stoica]{vllm_sosp23}
Kwon, W., Li, Z., Zhuang, S., Sheng, Y., Zheng, L., Yu, C.~H., Gonzalez, J., Zhang, H., and Stoica, I.
\newblock Efficient memory management for large language model serving with pagedattention.
\newblock In \emph{Proceedings of the 29th Symposium on Operating Systems Principles}, pp.\  611--626, 2023.

\bibitem[Larimi et~al.(2020)Larimi, Salami, Unsal, Kestelman, Sarbazi-Azad, and Mutlu]{hbm_arxiv2021}
Larimi, S. S.~N., Salami, B., Unsal, O.~S., Kestelman, A.~C., Sarbazi-Azad, H., and Mutlu, O.
\newblock Understanding power consumption and reliability of high-bandwidth memory with voltage underscaling, 2020.
\newblock URL \url{https://arxiv.org/abs/2101.00969}.

\bibitem[Liu et~al.(2024{\natexlab{a}})Liu, Wu, Chung, Lai, Lee, and Chowdhury]{andes_arxiv2024}
Liu, J., Wu, Z., Chung, J.-W., Lai, F., Lee, M., and Chowdhury, M.
\newblock Andes: Defining and enhancing quality-of-experience in llm-based text streaming services, 2024{\natexlab{a}}.
\newblock URL \url{https://arxiv.org/abs/2404.16283}.

\bibitem[Liu et~al.(2024{\natexlab{b}})Liu, Chen, Tian, Zou, Chen, and Cui]{memory_arxiv2024}
Liu, N., Chen, L., Tian, X., Zou, W., Chen, K., and Cui, M.
\newblock From llm to conversational agent: A memory enhanced architecture with fine-tuning of large language models, 2024{\natexlab{b}}.
\newblock URL \url{https://arxiv.org/abs/2401.02777}.

\bibitem[Liu et~al.(2024{\natexlab{c}})Liu, Li, Cheng, Ray, Huang, Zhang, Du, Yao, Lu, Ananthanarayanan, et~al.]{cachegen_sigcomm24}
Liu, Y., Li, H., Cheng, Y., Ray, S., Huang, Y., Zhang, Q., Du, K., Yao, J., Lu, S., Ananthanarayanan, G., et~al.
\newblock Cachegen: Kv cache compression and streaming for fast large language model serving.
\newblock In \emph{Proceedings of the ACM SIGCOMM 2024 Conference}, pp.\  38--56, 2024{\natexlab{c}}.

\bibitem[Minaee et~al.(2024)Minaee, Mikolov, Nikzad, Chenaghlu, Socher, Amatriain, and Gao]{large_arxiv2024}
Minaee, S., Mikolov, T., Nikzad, N., Chenaghlu, M., Socher, R., Amatriain, X., and Gao, J.
\newblock Large language models: A survey.
\newblock \emph{arXiv preprint arXiv:2402.06196}, 2024.

\bibitem[ModelTC(2024)]{lightllm_2024}
ModelTC.
\newblock Lightllm: A lightweight framework for large language model inference.
\newblock \url{https://github.com/ModelTC/lightllm}, 2024.
\newblock A Python-based LLM inference and serving framework with lightweight design, easy scalability, and high-speed performance.

\bibitem[Nijkamp et~al.(2023)Nijkamp, Pang, Hayashi, Tu, Wang, Zhou, Savarese, and Xiong]{codegen_arxiv2023}
Nijkamp, E., Pang, B., Hayashi, H., Tu, L., Wang, H., Zhou, Y., Savarese, S., and Xiong, C.
\newblock Codegen: An open large language model for code with multi-turn program synthesis, 2023.
\newblock URL \url{https://arxiv.org/abs/2203.13474}.

\bibitem[{NVIDIA}(2024)]{tensorrt_llm}
{NVIDIA}.
\newblock Nvidia tensorrt-llm.
\newblock \url{https://docs.nvidia.com/tensorrt-llm/index.html}, 2024.
\newblock Accessed: 2024-10-28.

\bibitem[{OpenAI}(2024)]{chatgpt}
{OpenAI}.
\newblock Chatgpt.
\newblock \url{https://openai.com/chatgpt}, 2024.
\newblock Accessed: 2024-10-28.

\bibitem[Patke et~al.(2024)Patke, Reddy, Jha, Qiu, Pinto, Cui, Narayanaswami, Kalbarczyk, and Iyer]{qlm_arxiv2024}
Patke, A., Reddy, D., Jha, S., Qiu, H., Pinto, C., Cui, S., Narayanaswami, C., Kalbarczyk, Z., and Iyer, R.
\newblock One queue is all you need: Resolving head-of-line blocking in large language model serving.
\newblock \emph{arXiv preprint arXiv:2407.00047}, 2024.

\bibitem[Qin et~al.(2024)Qin, Li, He, Zhang, Wu, Zheng, and Xu]{mooncake_arxiv2024}
Qin, R., Li, Z., He, W., Zhang, M., Wu, Y., Zheng, W., and Xu, X.
\newblock Mooncake: Kimi's kvcache-centric architecture for llm serving.
\newblock \emph{arXiv preprint arXiv:2407.00079}, 2024.

\bibitem[{ShareGPT}(2024)]{sharegpt}
{ShareGPT}.
\newblock Sharegpt: Share your wildest chatgpt conversations with one click.
\newblock \url{https://sharegpt.com/}, 2024.

\bibitem[Shaw et~al.(2018)Shaw, Uszkoreit, and Vaswani]{self_arxiv2018}
Shaw, P., Uszkoreit, J., and Vaswani, A.
\newblock Self-attention with relative position representations.
\newblock \emph{arXiv preprint arXiv:1803.02155}, 2018.

\bibitem[Sheng et~al.(2023)Sheng, Zheng, Yuan, Li, Ryabinin, Chen, Liang, R{\'e}, Stoica, and Zhang]{flexgen_icml23}
Sheng, Y., Zheng, L., Yuan, B., Li, Z., Ryabinin, M., Chen, B., Liang, P., R{\'e}, C., Stoica, I., and Zhang, C.
\newblock Flexgen: High-throughput generative inference of large language models with a single gpu.
\newblock In \emph{International Conference on Machine Learning}, pp.\  31094--31116. PMLR, 2023.

\bibitem[Sheng et~al.(2024)Sheng, Cao, Li, Zhu, Li, Zhuo, Gonzalez, and Stoica]{fairness_osdi24}
Sheng, Y., Cao, S., Li, D., Zhu, B., Li, Z., Zhuo, D., Gonzalez, J.~E., and Stoica, I.
\newblock Fairness in serving large language models.
\newblock In \emph{18th USENIX Symposium on Operating Systems Design and Implementation (OSDI 24)}, pp.\  965--988, 2024.

\bibitem[Sun et~al.(2024)Sun, Huang, Zhao, Xiao, Zhang, Li, and Lin]{llumnix_arxiv2024}
Sun, B., Huang, Z., Zhao, H., Xiao, W., Zhang, X., Li, Y., and Lin, W.
\newblock Llumnix: Dynamic scheduling for large language model serving, 2024.
\newblock URL \url{https://arxiv.org/abs/2406.03243}.

\bibitem[Touvron et~al.(2023)Touvron, Lavril, Izacard, Martinet, Lachaux, Lacroix, Rozi{\`e}re, Goyal, Hambro, Azhar, et~al.]{llama_arxiv2023}
Touvron, H., Lavril, T., Izacard, G., Martinet, X., Lachaux, M.-A., Lacroix, T., Rozi{\`e}re, B., Goyal, N., Hambro, E., Azhar, F., et~al.
\newblock Llama: Open and efficient foundation language models.
\newblock \emph{arXiv preprint arXiv:2302.13971}, 2023.

\bibitem[Von~Puttkamer(1975)]{buddy_1975}
Von~Puttkamer, E.
\newblock A simple hardware buddy system memory allocator.
\newblock \emph{IEEE Transactions on Computers}, C-24\penalty0 (10):\penalty0 953--957, 1975.
\newblock \doi{10.1109/T-C.1975.224100}.

\bibitem[Wei et~al.(2022)Wei, Wang, Schuurmans, Bosma, Xia, Chi, Le, Zhou, et~al.]{chain_arxiv2022}
Wei, J., Wang, X., Schuurmans, D., Bosma, M., Xia, F., Chi, E., Le, Q.~V., Zhou, D., et~al.
\newblock Chain-of-thought prompting elicits reasoning in large language models.
\newblock \emph{Advances in neural information processing systems}, 35:\penalty0 24824--24837, 2022.

\bibitem[Wu et~al.(2023)Wu, Zhong, Zhang, Huang, Liu, and Jin]{fastserve_arxiv23}
Wu, B., Zhong, Y., Zhang, Z., Huang, G., Liu, X., and Jin, X.
\newblock Fast distributed inference serving for large language models.
\newblock \emph{arXiv preprint arXiv:2305.05920}, 2023.

\bibitem[Yang et~al.(2024)Yang, Yang, Hui, Zheng, Yu, Zhou, Li, Li, Liu, Huang, et~al.]{qwen2_arxiv2024}
Yang, A., Yang, B., Hui, B., Zheng, B., Yu, B., Zhou, C., Li, C., Li, C., Liu, D., Huang, F., et~al.
\newblock Qwen2 technical report.
\newblock \emph{arXiv preprint arXiv:2407.10671}, 2024.

\bibitem[Yin et~al.(2024)Yin, Xu, Li, and Liu]{llmaas_arxiv2024}
Yin, W., Xu, M., Li, Y., and Liu, X.
\newblock Llm as a system service on mobile devices.
\newblock \emph{arXiv preprint arXiv:2403.11805}, 2024.

\bibitem[Zheng et~al.(2023)Zheng, Yin, Xie, Huang, Sun, Hao~Yu, Cao, Kozyrakis, Stoica, Gonzalez, et~al.]{sglang_arxiv2023}
Zheng, L., Yin, L., Xie, Z., Huang, J., Sun, C., Hao~Yu, C., Cao, S., Kozyrakis, C., Stoica, I., Gonzalez, J.~E., et~al.
\newblock Efficiently programming large language models using sglang.
\newblock \emph{arXiv e-prints}, pp.\  arXiv--2312, 2023.

\bibitem[Zhong et~al.(2024)Zhong, Liu, Chen, Hu, Zhu, Liu, Jin, and Zhang]{dist_arxiv2024}
Zhong, Y., Liu, S., Chen, J., Hu, J., Zhu, Y., Liu, X., Jin, X., and Zhang, H.
\newblock Distserve: Disaggregating prefill and decoding for goodput-optimized large language model serving, 2024.
\newblock URL \url{https://arxiv.org/abs/2401.09670}.

\bibitem[Zhu et~al.(2024)Zhu, Liu, Dong, Xu, Huang, Kong, Chen, and Li]{translation_arxiv2023}
Zhu, W., Liu, H., Dong, Q., Xu, J., Huang, S., Kong, L., Chen, J., and Li, L.
\newblock Multilingual machine translation with large language models: Empirical results and analysis, 2024.
\newblock URL \url{https://arxiv.org/abs/2304.04675}.

\end{thebibliography}
